%% file: emnlp2020.tex
\documentclass[11pt,a4paper]{article}
\RequirePackage[hyphens]{url}
\usepackage[hyperref]{emnlp2020}
\usepackage{times}
\usepackage{latexsym}
\usepackage{enumitem,kantlipsum}
\usepackage{textcomp}
\usepackage{subcaption}
\usepackage{booktabs}
\usepackage{comment}

\usepackage{algorithm}
\usepackage{algorithmic}
\usepackage{url}
\usepackage{todonotes}
\usepackage{color,soul}
\usepackage{graphicx}
\usepackage{multirow}
\usepackage{amsmath}
\usepackage[normalem]{ulem}

\input{math_commands.tex}

\usepackage{microtype}

\aclfinalcopy % Uncomment this line for the final submission
\title{Analyzing Redundancy in Pretrained Transformer Models}
\author{
	Fahim Dalvi ~~~ Hassan Sajjad ~~~ Nadir Durrani ~~~ Yonatan Belinkov\textsuperscript{*} \\ 
	{\tt \{faimaduddin,hsajjad,ndurrani\}@hbku.edu.qa} \\ 
	Qatar Computing Research Institute, HBKU Research Complex, Doha 5825, Qatar \\\\ 
	\textsuperscript{*}MIT Computer Science and Artificial Intelligence Laboratory and Harvard \\  
	John A.\ Paulson School of Engineering and Applied Sciences, Cambridge, MA, USA \\ 
	{\tt belinkov@csail.mit.edu}
}

\date{}

\begin{document}
	\maketitle
	\begin{abstract}
		Transformer-based deep NLP models are trained using hundreds of millions of parameters, limiting their applicability in computationally constrained environments. In this paper, we study the cause of these limitations by defining a notion of \emph{Redundancy}, which we categorize into two classes:  \emph{General Redundancy} and \emph{Task-specific Redundancy}. We dissect two popular pretrained models, BERT and XLNet, studying how much redundancy they exhibit at a representation-level and at a more fine-grained neuron-level. Our analysis reveals interesting insights, such as:  i) 85\% of the neurons across the network are redundant and ii) at least 92\% of them can be removed when optimizing towards a downstream task. Based on our analysis, we present an efficient feature-based transfer learning procedure,  which maintains 97\% performance while using at-most 10\% of the original neurons.\footnote{The code for the experiments in this paper is available at \url{https://github.com/fdalvi/analyzing-redundancy-in-pretrained-transformer-models}}
	\end{abstract}
	
	\section{Introduction}
	\label{sec:intro}

	Large pretrained models have improved the state- of-the-art in a variety of NLP tasks, with each new model introducing deeper and wider architectures causing a significant increase in the number of parameters. For example, BERT large \cite{devlin-etal-2019-bert}, NVIDIA's Megatron model,
	%\footnote{\url{https://nv-adlr.github.io/MegatronLM}}
	and Google's T5 model \cite{2019t5}	were trained using 340 million, 8.3 billion and 11 billion parameters respectively. 
	
	%However, a
	An emerging body of work shows that these models are over-parameterized and do not require all the representational power lent by the rich architectural choices during inference. For example, these models can be distilled \cite{sanh2019distilbert,sun-etal-2019-patient} or pruned \cite{voita-etal-2019-analyzing,Sajjad2020PoorMB}, with a minor drop in performance. Recent research ~\cite{MuBV17,ethayarajh-2019-contextual} analyzed contextualized embeddings in pretrained models and showed that the representations learned within these models are highly anisotropic. While these approaches successfully exploited over-parameterization and redundancy in pretrained models, the choice of what to prune is empirically motivated and the work does not directly explore the redundancy in the network. Identifying and analyzing redundant parts of the network is useful in: i) developing a better understanding of these models, ii) guiding research on compact and efficient models, and iii) leading towards better architectural choices.

	In this paper, we analyze redundancy in pretrained models. 	We classify it into
	\textit{general redundancy} and \textit{task-specific redundancy}. The former is defined as the redundant information present in a pretrained model irrespective of any downstream task. This redundancy is an artifact of over-parameterization and other training choices that force various parts of the models to learn similar information. The latter is motivated by pretrained models being universal feature extractors. We hypothesize that several parts of the network are specifically redundant for a given downstream task. 

	We study both \emph{general} and \emph{task-specific} redundancies at the \textit{representation-level} and at a more fine-grained \textit{neuron-level}.	Such an analysis allows us to answer the following questions: i) how redundant are the layers within a model? ii) do all the layers add significantly diverse information? iii) do the dimensions within a hidden layer represent different facets of knowledge, or are some neurons largely redundant? iv) how much information in a pretrained model is necessary for specific downstream tasks? and v) can we exploit redundancy to enable efficiency?

	We introduce several methods to analyze redundancy in the network.	Specifically, for general redundancy, we use \textit{Center Kernel Alignment}~\cite{pmlr-v97-kornblith19a} for layer-level analysis, and \textit{Correlation Clustering} for neuron-level analysis. For task-specific redundancy, we use \textit{Linear Probing}~\cite{shi-etal-2016-neural,belinkov:2017:acl} to identify redundant layers, and \textit{Linguistic Correlation Analysis}~\cite{dalvi:2019:AAAI} to examine	neuron-level redundancy.
	
	We conduct our study on two pretrained language models, BERT \cite{devlin-etal-2019-bert} and XLNet \cite{yang2019xlnet}. While these networks are similar in the number of parameters, they are trained using different training objectives, which accounts for interesting comparative analysis between these models.	For task-specific analysis, we present our results across a wide suite of downstream tasks: four core NLP sequence labeling tasks and seven sequence classification tasks from the GLUE benchmark~\cite{wang-etal-2018-glue}. Our analysis yields the following insights:
	\\~\\
	\noindent \textit{General Redundancy:}

	\begin{itemize}[leftmargin=*,parsep=2pt,topsep=2pt]
		\setlength\itemsep{-0.03em}
		\item Adjacent layers are most redundant in the network, with lower layers having greater redundancy with adjacent layers.
		\item Up to $85\%$ of the neurons across the network are redundant in general, and can be pruned to substantially reduce the number of parameters.
		\item Up to $94\%$ of neuron-level redundancy is exhibited within the same or neighbouring layers. 
	\end{itemize}

	\noindent \textit{Task-specific Redundancy:}

	\begin{itemize}[leftmargin=*,parsep=2pt,topsep=2pt]
		\setlength\itemsep{-0.03em}
		\item Layers in a network are more redundant w.r.t. core language tasks such as learning morphology as compared to sequence-level tasks.
		\item At least $92\%$ of the neurons are redundant with respect to a downstream task and can be pruned without any loss in task-specific performance.
		\item Comparing models, XLNet is more redundant than BERT.
		\item Our analysis guides research in model distillation and suggests preserving knowledge of lower layers and aggressive pruning of higher-layers.
	\end{itemize}

	Finally, motivated by our analysis, we present an \textbf{efficient feature-based transfer learning procedure} 	that exploits various types of redundancy present in the network. We first target layer-level task-specific redundancy using linear probes and reduce the number of layers required in a forward pass to extract the contextualized embeddings. We then filter out general redundant neurons present in the contextualized embeddings using  Correlation Clustering. Lastly, we remove task-specific redundant neurons using Linguistic Correlation Analysis. We show that one can reduce the feature set %(neurons) 
	to less than 100 neurons for several tasks while maintaining more than 97\% of the performance. Our procedure achieves a speedup of up to 6.2x in computation time for sequence labeling tasks.

	\section{Related Work}
	\label{sec:related}

	A number of studies have analyzed representations at %\hs{nadir you may have more citations on analyzing pretrained M}
	layer-level~\cite{conneau2018you, liu-etal-2019-linguistic,tenney2019learn, kim2020pretrained, belinkov-etal-2020-analysis} and at neuron-level~\cite{bau:2019:ICLR,dalvi:2019:AAAI,suau2020finding, durrani-2020-individualNeurons}. These studies aim at analyzing either the linguistic knowledge learned in representations and in neurons or the general importance of neurons in the model. The former is commonly done using a probing classifier~\cite{shi-etal-2016-neural, belinkov:2017:acl,hupkes2018visualisation}. Recently,~\newcite{voita2020informationtheoretic,pimentel2020informationtheoretic} proposed probing methods based on information theoretic measures. %similarity-based methods  have been introduced %One of the common approach following by  
	The general importance of neurons is mainly captured using similarity and correlation-based methods~\cite{NIPS2017_7188_svcca,chrupala-alishahi-2019-correlating,Wu2020SimilarityAO}.
	Similar to the work on analyzing deep NLP models, we analyze pretrained models at representation-level and at neuron-level. Different from them, we analyze various forms of redundancy in these models. We draw upon various techniques from the literature and adapt them to perform a redundancy analysis. 
	
	While the work on pretrained model compression~\cite{cao2019,shen2019qbert,sanh2019distilbert,turc2019wellread,Gordon2019CompressingBS,Guyon:jmlr} indirectly shows that models exhibit redundancy, % information present in the network is redundant, 
	little has been done to explore the redundancy in the network. Recent studies~\cite{voita-etal-2019-analyzing,michel2019sixteen,Sajjad2020PoorMB,fan2019reducing} dropped attention heads and layers in the network with marginal degradation in performance. Their work is limited in the context of redundancy as none of the pruning choices are built upon the amount of redundancy present in different parts of the network. Our work identifies redundancy at various levels of the network and can guide the research in model compression.

	\section{Experimental Setup}
	\label{sec:experimental-setup}

	\subsection{Datasets and Tasks}
	
	% To analyze the general redundancy in pre-trained models, we use two broad categories of downstream tasks -- Sequence Labeling tasks and Sequence Classification Tasks. For the sequence labeling tasks, we study core linguistic tasks, i) part-of-speech (POS) tagging, 
	% ii) CCG super tagging, iii) semantic tagging (SEM) using and iv) syntactic chunking. For the sequence classification tasks, we study tasks from the GLUE benchmark~\citep{wang-etal-2018-glue}.\footnote{We did not evaluate on CoLA and WNLI because of the irregularities in the data and instability during the fine-tuning process: \url{https://gluebenchmark.com/faq}.}

	To analyze the general redundancy in pre-trained models, we use the Penn Treebank development set~\cite{marcus-etal-1993-building}, which consists of roughly 44,000 tokens. 
	For task-specific analysis, we use two broad categories of downstream tasks -- \emph{Sequence Labeling} and \emph{Sequence Classification} tasks. 
	For the sequence labeling tasks, we study core linguistic tasks, i) part-of-speech (POS) tagging using the 
	Penn TreeBank, %~\cite{marcus-etal-1993-building}, 
	ii) CCG super tagging using CCGBank~\cite{hockenmaier2006creating}, iii) semantic tagging (SEM) using Parallel Meaning Bank data~\cite{AbzianidzeBos2017IWCS} and iv) syntactic chunking using CoNLL 2000 shared task dataset~\cite{tjong-kim-sang-buchholz-2000-introduction}.
	
	For sequence classification, we study tasks from the GLUE benchmark~\citep{wang-etal-2018-glue}, namely i) sentiment analysis (SST-2) \cite{socher-etal-2013-recursive}, ii) semantic equivalence classification (MRPC) \cite{dolan-brockett-2005-automatically}, iii) natural language inference (MNLI) \cite{williams-etal-2018-broad}, iv) question-answering NLI (QNLI) \cite{rajpurkar-etal-2016-squad}, iv) question pair similarity\footnote{\url{http://data.quora.com/First-Quora-Dataset-Release-Question-Pairs}} (QQP), v) textual entailment (RTE) \cite{Bentivogli09thefifth}, and vi) semantic textual similarity \cite{cer-etal-2017-semeval}.\footnote{We did not evaluate on CoLA and WNLI because of the irregularities in the data and instability during the fine-tuning process: \url{https://gluebenchmark.com/faq}.} Complete statistics for all datasets is provided in Appendix \ref{subsec:appendix-data}.
	
	\paragraph{Other Settings} The neuron activations for each word 
	%$w_j$ 
	in our dataset are extracted from the pre-trained model for sequence labeling while the \texttt{[CLS]} token's representation (from a fine-tuned model) is used for sequence classification. The fine-tuning step is essential to 
	%For sequence classification tasks, in order to 
	optimize the \texttt{[CLS]} token for sentence representation. %, we fine-tune the models before conducting any analysis. 
	In the case of sub-words, 
	%the sub-word representations are aggregated by picking
	we pick the last sub-word's representation 
	%as found to be optimal for Neural MT %models
	~\cite{durrani-etal-2019-one,liu-etal-2019-linguistic}. 
	%and pre-trained Language Models ~\cite{liu-etal-2019-linguistic}.  %\cite{durrani-etal-2019-one}. %\yb{cite Nelson's paper for this practice} 
	%
	For sequence labeling tasks, we use training sets of 150K tokens, and standard development and test splits.
	%\nd{should we justify this by citing a paper that encourages lesser data for probing classifier}
	For sequence classification tasks, we set aside $5\%$ of the training data and use it to optimize all the parameters involved in the process and report results on development sets, since the test sets are not publicly available.

	\subsection{Models} We present our analysis 
	%and results 
	on two transformer-based pretrained
	models, BERT-base \cite{devlin-etal-2019-bert} and XLNet-base \cite{yang2019xlnet}.\footnote{We could not run BERT and XLNet large because of computational limitations. See the official BERT readme describing the issue \url{https://github.com/google-research/bert\#out-of-memory-issues}} The former is a masked language model, 
	%based on an auto-encoding objective, 
	while the latter is of an auto-regressive nature. 
	%\sout{This fundamental difference in %their 
	%design %can 
	%leads to interesting comparative analysis.} 
	% We use the base (12-layer) models for %both of these networks.
	% our experiments.\footnote{We could not run BERT and XLNet large because of computational limitations. See the official BERT readme describing the issue \url{https://github.com/google-research/bert\#out-of-memory-issues}} %We use un-finetuned models for the general redundancy analysis, as well as the experiments on sequence labeling task (since they represent core NLP tasks that should be learned by a general language model \hl{cite?}) \nd{I think this is unnecessary }
	%For the experiments on sequence classification tasks, we %first 
	%finetune the pre-trained models towards each task before conducting our analysis.\footnote{This is not required for sequence labeling task since they represent core linguistic properties (morphology, syntax, semantics) learned by general language model} 
	We use the \texttt{transformers} library \cite{Wolf2019HuggingFacesTS} to fine-tune these 
	%pre-trained 
	models using 
	%their 
	default hyperparameters.% settings.
	
	\paragraph{Classifier Settings} %We use a logistic regression classifier trained with ElasticNet regularization trained on the neuron representations as our probe for all task-specific analysis. 
	For layer-level probing and neuron-level ranking, we use a logistic regression classifier with ElasticNet regularization.
	%(as prescribed in \newcite{dalvi:2019:AAAI}), to extract neuron ranking.
	%This classifier is trained using the same settings as defined by \cite{dalvi:2019:AAAI}, in order to follow their procedure for neuron level analysis. 
	%Specifically, w
	We train the classifier for $10$ epochs with a learning rate of $1e^{-3}$, batch size of $128$ and a value of $1e^{-5}$ for both $L1$ and $L2$ lambda regularization parameters. 
	%We consider a relative loss of $3\%$ w.r.t. to the original performance as our threshold for acceptable degradation in performance in all of our analysis.
	%\nd{the choice of 3\% is not well motivated in this new story now.}

	\section{Problem Definition}
	\label{sec:redundancy-analysis}
	%\yb{how about a 2x2 table for these?}
	% hs: would be good if space permits
	%\nd{Merge Sections 4 and 5? Make 5 5.2 and 6 5.3 and 5.2?}
	Consider a pretrained 
	%neural network 
	model $\mathbf{M}$ with $L$ layers: $\{l_0, l_1, \ldots, l_L\}$, where $l_0$ is an embedding layer and each layer $l_i$ is of size $H$. Given a dataset $ \sD=\{w_1, w_2, ..., w_T\}$ consisting of $T$ words, the contextualized embedding of word $w_j$ at layer $l_i$ is $\vz_j^i = l_i(w_j)$. A \textit{neuron} consists of each individual unit of $\vz_j^i$. For example, BERT-base has $L=13$ layers,
	each 
	%including the embedding layer and each layer is 
	of size 768 i.e. there are 768 individual neurons in each layer. The total number of neurons in the model are $13\times768=9984$.
	
	We analyze redundancy in $\mathbf{M}$ at layer-level $l_i$: \textit{how redundant is a layer?} and at neuron-level: \textit{how redundant are the neurons?} We target these two questions in the context of general redundancy and task-specific redundancy. 
	
	\paragraph{Notion of redundancy:}  We broadly define redundancy to cover a range of observations. For example, we imply high similarity as a reflection of redundancy.  Similarly, for task-specific neuron-level redundancy, we hypothesize that some neurons additionally might be irrelevant for the downstream task in hand. There, we consider irrelevancy as part of the
	%the neuron-level 
	redundancy
	analysis. Succinctly, two neurons are considered to be redundant if they serve the same purpose from the perspective of feature-based transfer learning for a downstream task. 
	%to analyze the minimum information needed to perform a downstream task.
	%\nd{I would say again this is creating more confusion as it is only related to task specific redundancy. This becomes clear in the sections below}   
	
	% present four different analysis methods, each targeting a specific kind of redundancy in a given pre-trained language model. 
	%\nd{this notation was not used during this section much and can be shortened or removed?}
	
	\section{General Redundancy}
	\label{sec:general-redundancy-analysis}
	Neural networks are designed to be distributed in nature and are therefore innately redundant. Additionally, over-parameterization in pretrained models with a combination of various training and design choices 
	%such as dropouts 
	causes further redundancy of information. In the following, we 
	%present methods to 
	analyze general redundancy at layer-level and at neuron-level.
	
	\begin{figure}[t]
		\begin{subfigure}{.49\linewidth}
			\centering
			\includegraphics[width=\linewidth]{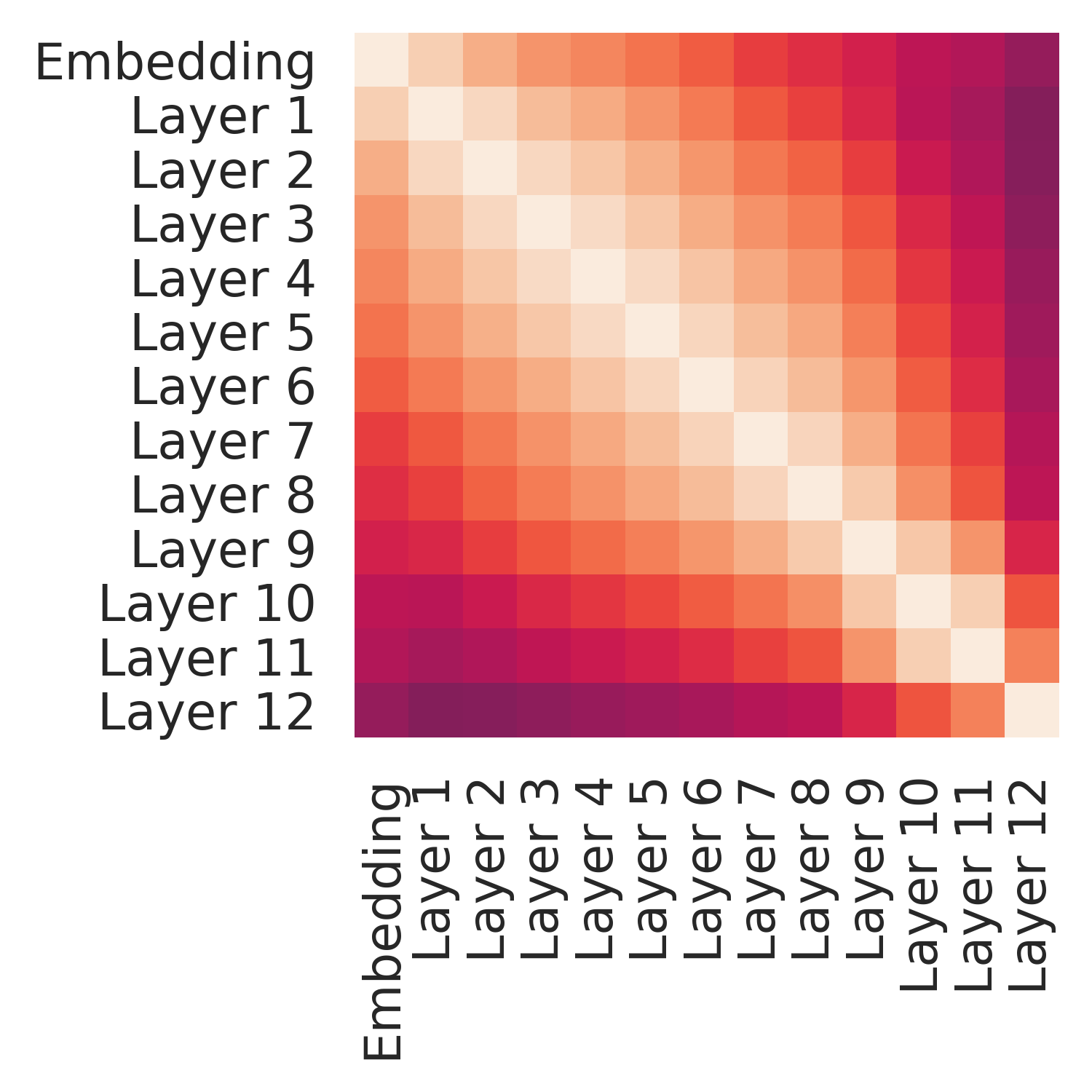}  
			\caption{BERT}
			\label{subfig:bert-lincka}
		\end{subfigure}
		\begin{subfigure}{.49\linewidth}
			\centering
			\includegraphics[width=\linewidth]{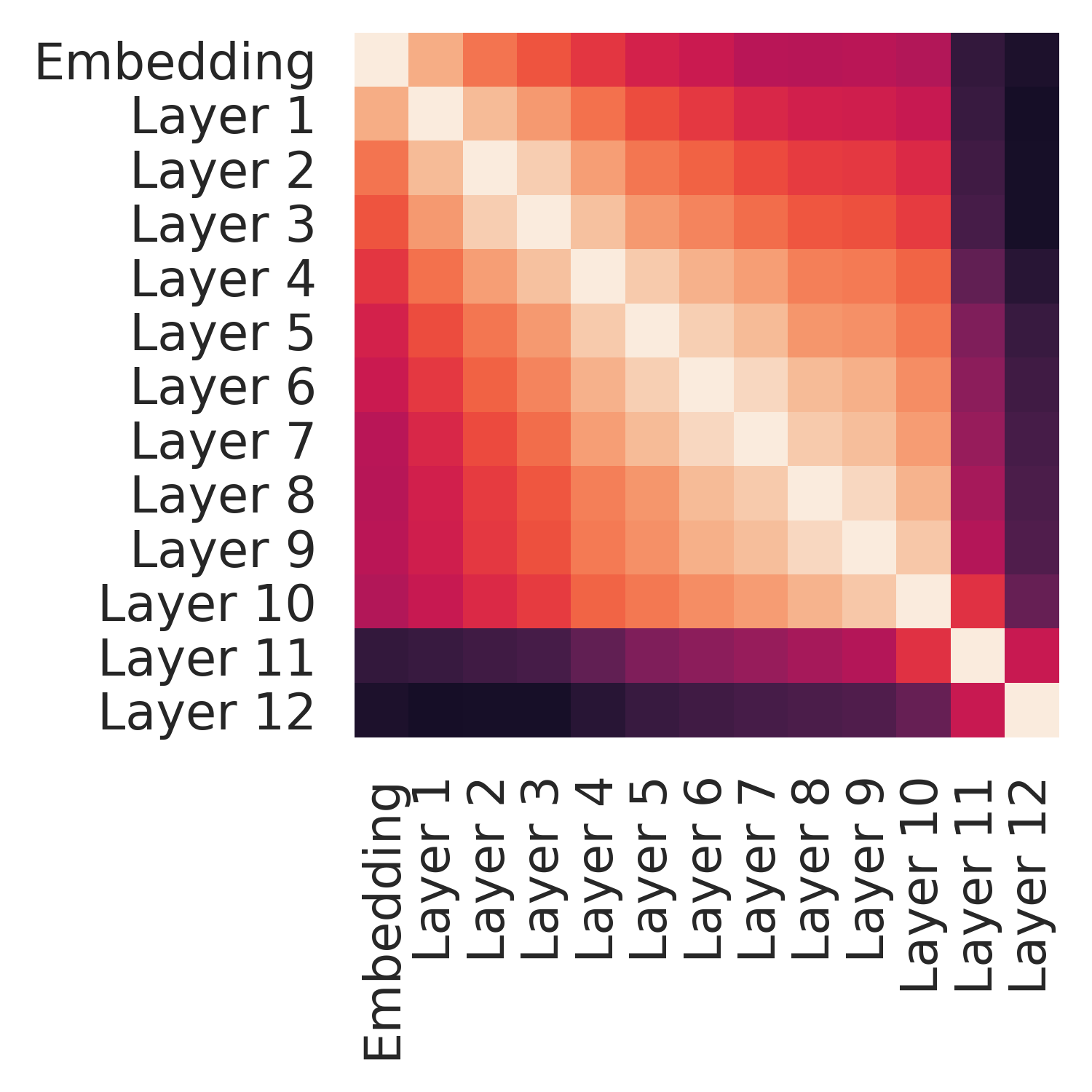}
			\caption{XLNet}
			\label{subfig:xlnet-lincka}
		\end{subfigure}
		\caption{ Pairwise 
			Similarity between 
			%all 
			the layers. % in the pretrained models.
			Brighter colors indicate higher similarity.}
		\label{fig:lincka}
	\end{figure}

	\subsection{Layer-level Redundancy}
	\label{subsec:layer-general-redundancy}
	
	We compute layer-level redundancy by comparing representations from different layers in a given model using linear Center Kernel Alignment (\texttt{cka} - \newcite{pmlr-v97-kornblith19a}). %The 
	\texttt{cka} %similarity 
	%measure brings
	is invariant to isotropic similarity and orthogonal transformation. In other words, the similarity measure itself does not depend on the various representations having neurons or dimensions with exactly the same distributions, but rather assigns a high similarity if the two representations behave similarly over \emph{all} the neurons. Moreover, \texttt{cka} is known to outperform other methods such as  CCA~\cite{Andrew_deepcanonical} and SVCCA~\cite{NIPS2017_7188_svcca}, 
	%and  linear regression, 
	in identifying relationships between different layers across different architectures.
	%
	%\yb{this paragraph is copied/redundant with the previous one} 
	%For analyzing layer-level redundancy, 
	% Our goal is to compare representations from various layers in a given model with each other. \newcite{pmlr-v97-kornblith19a} introduced linear centered kernel alignment (\texttt{cka}) as a method to calculate the similarity between representations of different neural networks or within the same network. The advantage \texttt{cka} brings is in-variance to isotropic similarity and orthogonal transformation. In other words, the similarity measure itself does not depend on the various representations having neurons or dimensions with exactly the same distributions, but rather assigns a high similarity if the two representations behave similarly over \emph{all} of their neurons. Compared to other similarity methods such as CCA~\cite{Andrew_deepcanonical}, SVCCA~\cite{NIPS2017_7188_svcca}, linear regression, etc., \texttt{cka} performed consistently well in identifying relationship between different layers across different architectures~\cite{pmlr-v97-kornblith19a}. 
	%
	%We consider \texttt{cka} to compare the layer representations within a pre-trained model. 
	%compared \texttt{cka} t is The similarity measure itself does not depend on the various representations having neurons or dimensions with exactly the same distributions, but rather assigns a high similarity if the two representations behave similarly over \emph{all} of their neurons. 
	While there are several other methods proposed in literature to analyze and compare representations \citep{10.3389/neuro.06.004.2008,bouchacourt-baroni-2018-agents,chrupala-alishahi-2019-correlating,chrupala-2019-symbolic}, we do not intend to compare them here and instead use \texttt{cka} to show redundancy in the network. The mathematical definition of \texttt{cka} is provided in Appendix \ref{subsec:appendix-cka} for the reader.
	
	We compute pairwise similarity between all %the 
	$L$ layers in the pretrained model and show the corresponding heatmaps in Figure \ref{fig:lincka}.
	%for studying the general layer-level redundancy in the network. 
	We hypothesize that a high similarity entails (general) redundancy.
	%Figure \ref{fig:lincka} shows similarity heatmaps of BERT and XLNet.
	% according to \texttt{cka}. 
	%\sout{presents the similarities between all pairs of layers in the pre-trained models as heatmaps.}
	%\nd{the first sentence of paragraph or section should always summarize the paragraph. Can we start with the finding and then refer to results?} 
	%We can see that t
	Overall the \textbf{similarity between adjacent layers is high}, indicating that the change of encoded knowledge from one layer to another takes place in small incremental steps as we move from a lower layer to a higher layer. %The only layer pair 
	%An exception to where this is distinctively false 
	An exception to this observation is the final %layer pair
	pair of layers, %i.e. Layer
	$l_{11}$ and $l_{12}$, whose %where the 
	similarity is much lower than other adjacent pairs of layers. We %hypothesis 
	speculate that this is because the final layer is highly optimized for the objective at hand, while the lower layers try to encode as much general linguistic knowledge as possible. This has also been alluded to by others~\cite{hao-etal-2019-visualizing,Wu2020SimilarityAO}. %\yb{can cite here other papers showing the top layers change more during training; there are refs in our acl 2020 paper, I think}
	
	%\yb{how to move from similarity to redundnacy? maybe just say that we hypothesize that a high similarity entails (general) redundancy} 
	%\hs{i am not sure if this paragraph is needed or not}\nd{yes I would move it to related work}\yb{agree, but I do think it's good to acknowledge here that there may be other methods, and perhaps refer to the related work, or say that future work can study other methods} 
	
	\begin{figure}[t]
	\begin{subfigure}{\linewidth}
		\centering
		\includegraphics[width=.8\linewidth]{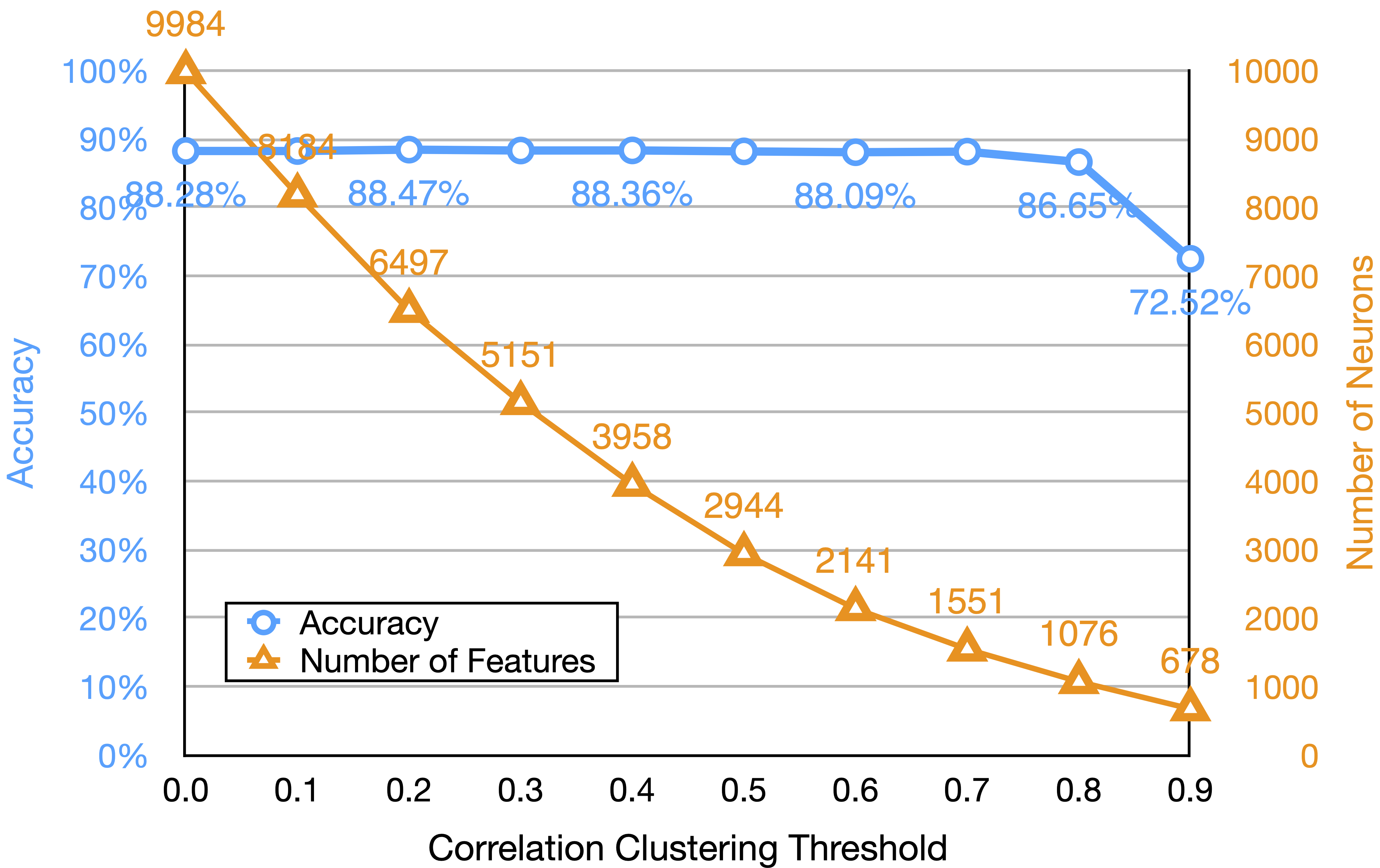}
		\caption{BERT}
		\label{subfig:bert-cc-reduction}
	\end{subfigure}
	\begin{subfigure}{\linewidth}
		\centering
		\includegraphics[width=.8\linewidth]{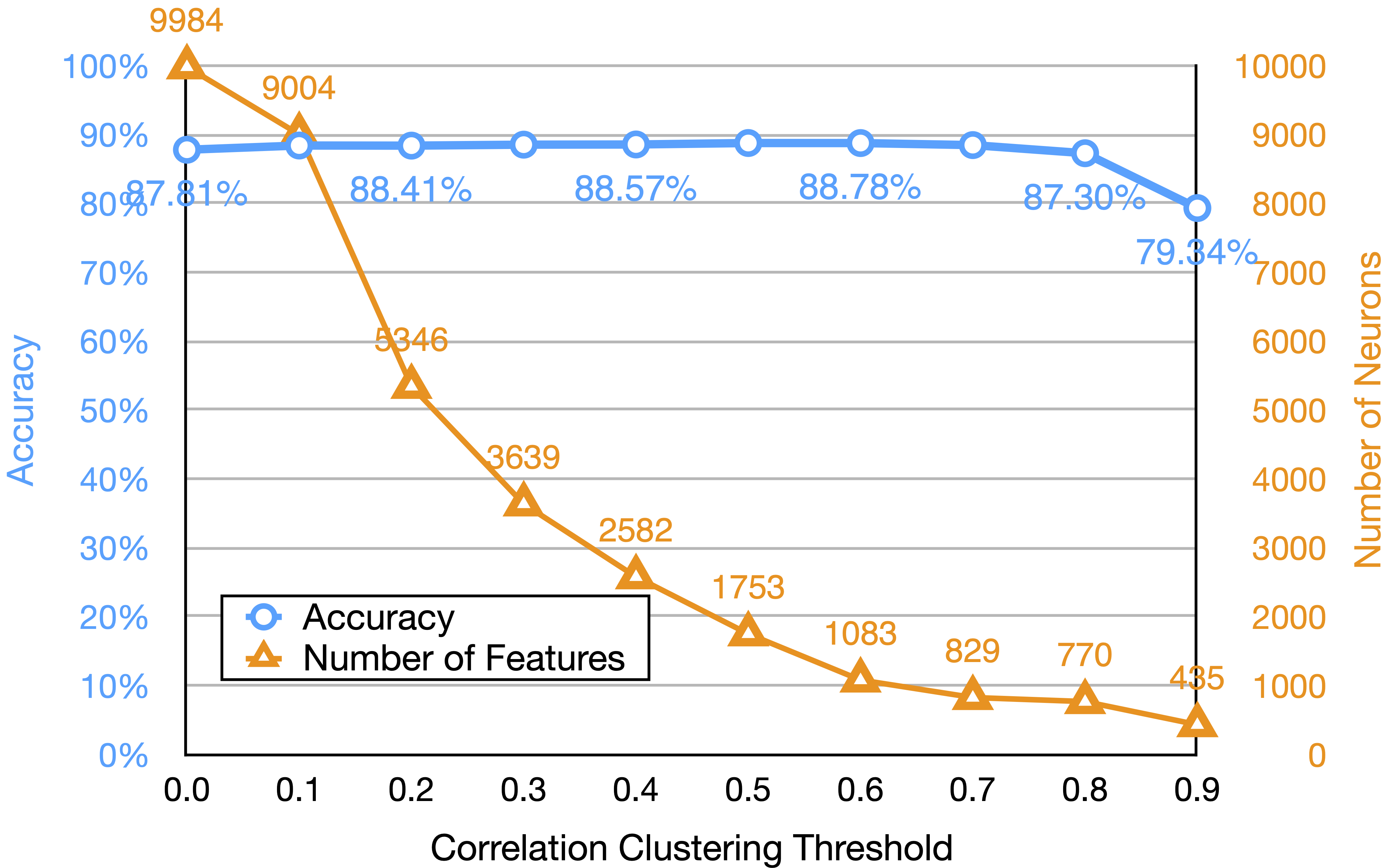}
		\caption{XLNet}
		\label{subfig:xlnet-cc-reduction}
	\end{subfigure}
		\caption{General neuron-level redundancy in BERT and XLNet; comparing the average reduction of neurons for different CC thresholds and the average accuracy across all downstream tasks. See Appendix \ref{subsec:appendix-general-neuron} for detailed per-task results. }
	\label{fig:cc-reduction}
\end{figure}
	
	\subsection{Neuron-level Redundancy}
	\label{subsec:neuron-general-redundancy}
	Assessing redundancy at the layer level may be too coarse grained. Even if a layer is not redundant with other layers, a subset of its neurons may still be redundant.
	%\sout{While it is important to see how layers in the same network are similar and redundant in the overall knowledge they encode, it is not reflective of the amount of 
	%how much 
	%redundancy exist at 
	%a fine-grained
	%neuron level. It is possible that a layer is not redundant to other layers, but several of %it's 
	%the underlying neurons are redundant across the network.} %\sout{it is also helpful to analyze redundancy at a fine-grained neuron level.} %finer level, i.e. at the neuron level.
	We analyze neuron-level redundancy in a network using correlation clustering -- \texttt{CC}~\cite{correlationclustering}. We group neurons with highly correlated activation patterns over all of the words $w_j$. Specifically, every neuron in the vector $\vz^i_j$ from some layer $i$ can be represented as a $T$ dimensional vector, where each index is the activation value $\vz^i_j$ of that neuron for some word $w_j$, where $j$ ranges from 1 to $T$. We calculate the Pearson product-moment correlation of every neuron vector $\vz^i$ with every other neuron. This results in a $N \times N$ matrix $corr$, where $N$ is the total number of neurons and $corr(x, y)$ represents the correlation between neurons $x$ and $y$. The correlation value ranges from $-1$ to $1$, giving us a relative scale to compare any two neurons. 
	%compute similarity or redundancy between any two neurons. We hypothesize that a high similarity value entails redundancy. 
	%\yb{again need to explicitly say/argue/assume that similarity entails redundancy} %If the absolute correlation value is high between two neurons, we can hypothesis that they encode very similar information and are thus redundant.
	A high absolute correlation value between two neurons implies that they encode very similar information and therefore are redundant.
	%We hypothesize that a high similarity value entails redundancy of information. 
	We convert 
	%the correlation matrix
	$corr$ into a distance matrix $cdist$ by applying $cdist(x,y) = 1 - |corr(x,y)|$ %. We 
	and cluster the distance matrix $cdist$ by using agglomerative hierarchical clustering with average linkage\footnote{We experimented with other clustering algorithms such as k-means and DBSCAN, and did not see any noticeable difference in the resulting clusters.} %which minimizes 
	to minimize the average distance of all data points in pairs of clusters. The maximum distance between any two points in a cluster is controlled by the hyperparameter $c_t$. It ranges from $0$ to $1$ where a high value results in large-sized clusters with a small number of total clusters. %\yb{this paragraph is very long and detailed; some stuff can be moved to appendix if we need space} % i.e. directly affects the size of the resulting clusters.

	\begin{figure}[t]
		\centering
		\includegraphics[width=0.8\linewidth]{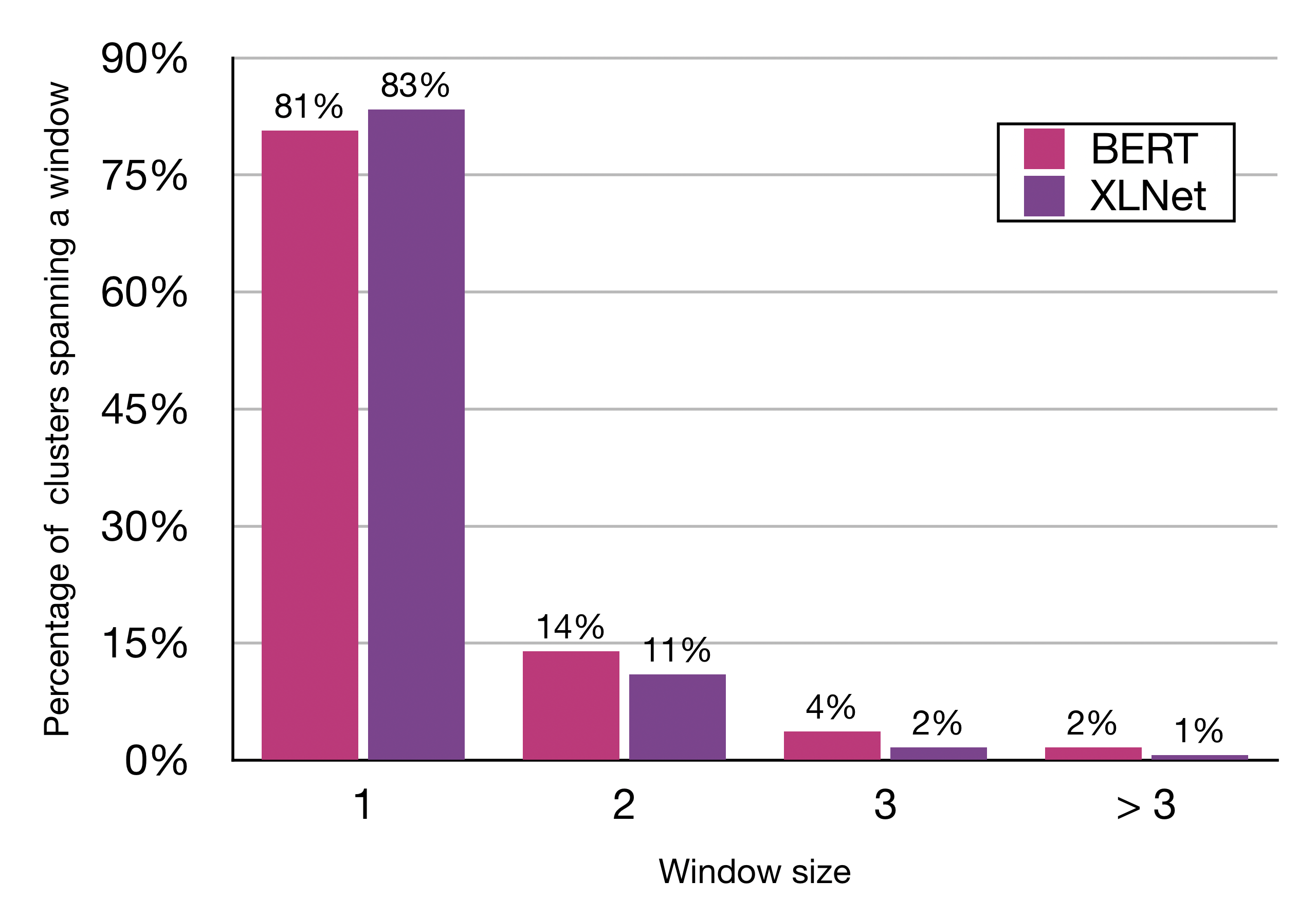}
		\caption{Percentage of clusters which contain neurons from the same layers, adjacent layers, within three neighboring layers and more than three layers apart.}
		\label{fig:cluster-distribution}
	\end{figure}
	
	\paragraph{Substantial amount of neurons are redundant}
	In order to evaluate the effect of clustering in combining redundant neurons, we randomly pick a neuron from each cluster and form a reduced set of non-redundant neurons. Recall that the clustering is applied independently on the data without using any task-specific labels. We then build task-specific classifiers for each task on the reduced set and analyze the average accuracy. If the average accuracy of a reduced set is close to that of the full set of neurons, we conclude that the reduced set has filtered out redundant neurons.
	%
	%Given a set of clusters, we randomly pick one neuron from each cluster to form a reduced set of neurons. of the task and task-specific classifiers are built on the reduced set of neurons later.
	%
	Figure \ref{fig:cc-reduction} shows the effect of clustering on BERT and XLNet using different values of $c_t$ with respect to average performance across all tasks.
	%Given a set of clusters, we randomly pick one neuron from each cluster to form a reduced set of neurons. The clustering is applied independent of the task and task-specific classifiers are built on the reduced set of neurons later.
	It is remarkable to observe that \textbf{85\% of neurons can be removed without any loss in accuracy} ($c_t=0.7$) in BERT, alluding to a high-level of neuron-level redundancy. 
	We observe an even higher reduction in XLNet. At $c_t=0.7$, 92\% of XLNet neurons can be removed while maintaining oracle performance.
	%XLNet showed similar trend. Due to space limitation, we moved it to appendix.
	%Comparing BERT and XLNet, the curve of XLNet is much steeper reflecting higher redundancy compared to BERT.
	%
	%Since our distances range from $0$ to $1$, with $0$ being highly correlated, we choose a value of $0.3$ for $c_t$. This value was chosen by searching through a range of thresholds between $0$ to $1$ and confirming no loss of performance across all tasks in our task set at the chosen value.
	%
	%We perform a qualitative analysis of the clusters created by \textit{correlation clustering} at $c_t=0.3$. %Firstly, w
	%Figure \ref{fig:sentence_heatmap} 
	We additionally visualize a few neurons within a cluster. The activation patterns are quite similar in their behavior, though not identical, highlighting the efficacy of \texttt{CC} in clustering neurons with analogous behavior. An activation heatmap for several neurons is provided in Appendix \ref{subsec:appendix-general-neuron}. %Hence, we tune the clustering algorithm to not cluster very aggressively (by setting the maximum distance threshold to a smaller value, in our case $0.3$).
	
	\paragraph{Higher neuron redundancy within and among neighboring layers} %\yb{Consider moving this paragraph before the previous one, because this talks about the clustering makeup in general, without any classification, while the previous one introduces classification (and so is no longer completely "general redundancy", btw)} 
	%\nd{We have always talked about layers before neurons even below. I see why you flipped the order here but I think you can move the like "This reflects ..." even in the other section}
	% we are talking about neurons here.
	
	We analyze the general makeup of the clusters at $c_t=0.3$.\footnote{The choice of $0.3$ avoids aggressive clustering and enables the analysis of the most redundant neurons.} Figure \ref{fig:cluster-distribution} shows the percentage of clusters that contain neurons from the same layer (window size 1), neighboring layers (window sizes 2 and 3) and from layers further apart. We can see that a vast majority of clusters ($\approx95\%$) either contain neurons from the same layer or from adjacent layers. This reflects that the main source of redundancy is among the individual representation units in the same layer or neighboring layers of the network. 
	The finding motivates pruning of models by compressing layers as oppose to reducing the overall depth in a distilled version of a model.
	%\nd{it would be great to mention that we found this observation to be very useful in our contemptuous work on pruning neural networks.} we did not compress layers instead of reduce depth in our other work
	
	% \begin{figure*}[ht]
	% 	\centering
	% 	\begin{subfigure}{0.45\linewidth}
	% 		\centering
	% 		\includegraphics[width=0.8\linewidth]{figs/layerwise-redundancy-bert.png}  
	% 		\caption{BERT}
	% 		\label{subfig:bert-summary-layerwise}
	% 	\end{subfigure}
	% 	\begin{subfigure}{.45\linewidth}
	% 		\centering
	% 		\includegraphics[width=0.8\linewidth]{figs/layerwise-redundancy-xlnet.png}  
	% 		\caption{XLNet}
	% 		\label{subfig:xlnet-summary-layerwise}
	% 	\end{subfigure}
	% 	\caption{Task-specific layer-wise redundant layers represented by the green blocks. %Green blocks represent task-specific redundant layers. % with respect to a task. %For each task, the figures show the number of the layers that have close to the optimal performance when compared to a classifier that uses features from all of the layers.
	% 	}
	% 	\label{fig:summary-layerwise}
	% \end{figure*}

	\section{Task-specific Redundancy}

	% \nd{the opening sentence is not clear}
	% While there exists general redundancy in pretrained models,
	% \sout{we hypothesize that} several of \sout{their} parts \sout{are} \hll{may be even} redundant 
	% %and irrelevant 
	% for a downstream task. 
	%may contain task-specific redundancies and have they are successfully applied to downstream tasks. 
	%may contain redundant and irrelevant information for a downstream task. 
	%Pre-trained models contain task-specific redundancies that extend over\yb{what does "extend over" mean here?} the general redundancies targeted by the previous methods. 
	%\sout{While the previous sections have shown a high amount of general redundancy in pretrained models, they may additionally exhibit redundancies specific to a downstream task.}
	%
	While pretrained models have a high amount of general redundancy as shown in the previous section, they may additionally exhibit redundancies specific to a downstream task.
	%
	%\hll{While pre-trained models have high amount of general redundancy as we have shown, they may be even redundant with respect to a downstream task.} 
	Studying redundancy in relation to a specific task helps us understand pretrained models better. It further reflects on how much of the network, and which parts of the network, suffice to perform a 
	%downstream 
	task efficiently.

	\subsection{Layer-level Redundancy} %\nd{this section is very dense and needs significant rewriting to present the results in a better way}
	\label{subsec:layer-task-specific-redundancy}
	
	%Pretrained models being a universal feature extractor when used for a downstream task, may contain redundant and less relevant information for the task. 
	%A pre-trained model can 
	%Pre-trained models contain task-specific redundancies that extend over the general redundancies targeted by the previous methods. 
	%\textcolor{red}{
	%\nd{this is not clear and also not intuitive. irrelevancy versus redundancy should be clarified} 
	%\sout{Note that the set of neurons redundant for a specific task also includes neurons irrelevant for that task, apart from neurons that are relevant but similar in the information they encode.}} 
	%in the context of transfer learning for the task
	%Studying the overall redundancy in relation to a specific task can not only help us to understand pre-trained models better, but can also inform us about how much of the original network is sufficient for efficiently solve the given task.
	%To analyze task-specific this, we use %a linear probe \cite{alain2016understanding, belinkov2019linguistic} 
	%probing classifiers \cite{shi-padhi-knight:2016:EMNLP2016, belinkov:2017:acl} trained on each layer $l_i$ %and the task %
	%
	%Hence, studying the overall redundancy with relation to a specific task can not only help us understand pre-trained models better in the context of transfer learning for the task, but can also inform us as to how much of the original network is sufficient for efficiently solve the given task. To analyze this, 
	
	To analyze layer-level task-specific redundancy, we train linear probing classifiers \cite{shi-padhi-knight:2016:EMNLP2016, belinkov:2017:acl} on each layer $l_i$ (\textit{layer-classifier}). We consider a classifier's performance as a proxy for the amount of task-specific knowledge learned by a layer. Linear classifiers are a popular choice in analyzing deep NLP models due to their better interpretability~\cite{qian-qiu-huang:2016:EMNLP2016,belinkov-etal-2020-analysis}. \newcite{hewitt-liang-2019-designing} have
	%also 
	shown linear probes to have higher \emph{Selectivity}, a property deemed desirable for more interpretable probes. 
	%\yb{here we may get criticism on using probing classifiers, given the recent concerns (like control tasks of Hewitt and the information-theoretic MDL of Voita). We can anticipate this by adding somewhere that probing classifiers have limitations (and give refs), and that perhaps limits the results. We can stress that we look at relative performances, compared to the oracle, because that mitigates some of the concerns of Pimentel; but it doesn't help with the other concerns, of Voita and Hewitt } % for the given task. 
	%If a classifier trained on a layer achieve similar performance to oracle, 
	%If the classifiers trained on two different layers achieve a similar performance, this implies they encode redundant knowledge with respect to the underlying task. \hll{Note that this does not mean that those layers are identical or they represent the knowledge about the task in a similar way. Instead, they both have enough knowledge about the task to perform it well.}
	%\yb{this is a very strong assumption IMO. two representations might lead to similar accuracies in different ways. Maybe acknowledge this limitation} 
	%
	%\hll{For any classifier trained using layer $l_i$, that performs close to oracle (within a performance threshold), we imply that $l_i$ encodes sufficiently close knowledge to oracle and is therefore redundant to any other layers that achieve oracle performance.} 

	We compare each \textit{layer-classifier} with an oracle-classifier trained 
	%\sout{by concatenating all layers}
	over concatenation of all layers of the network. For all individual layers that perform close to oracle (maintaining 99\% of the performance in our results), we imply that they encode sufficient knowledge about the task and are therefore redundant in this context.
	%\sout{The performance of all those layer-classifiers that are within a performance threshold of the oracle, we imply that all those layers encode enough knowledge about the task to perform it well. Thus they can be considered as redundant for the underlying task.} 
	% Note that this does not necessarily imply that those layers are identical or they represent the knowledge in a similar way -- instead they have enough knowledge about the task to perform it well.
	Note that this does not necessarily imply that those layers are identical or that they represent the knowledge in a similar way -- instead they have redundant overall knowledge specific to the task at hand. 
	%\nd{While it is important to say this, it is kind of weakening our notion of redundancy}
	
	\paragraph{High redundancy for core linguistic tasks}
	\begin{figure}[t]
		\centering
		\includegraphics[width=\linewidth]{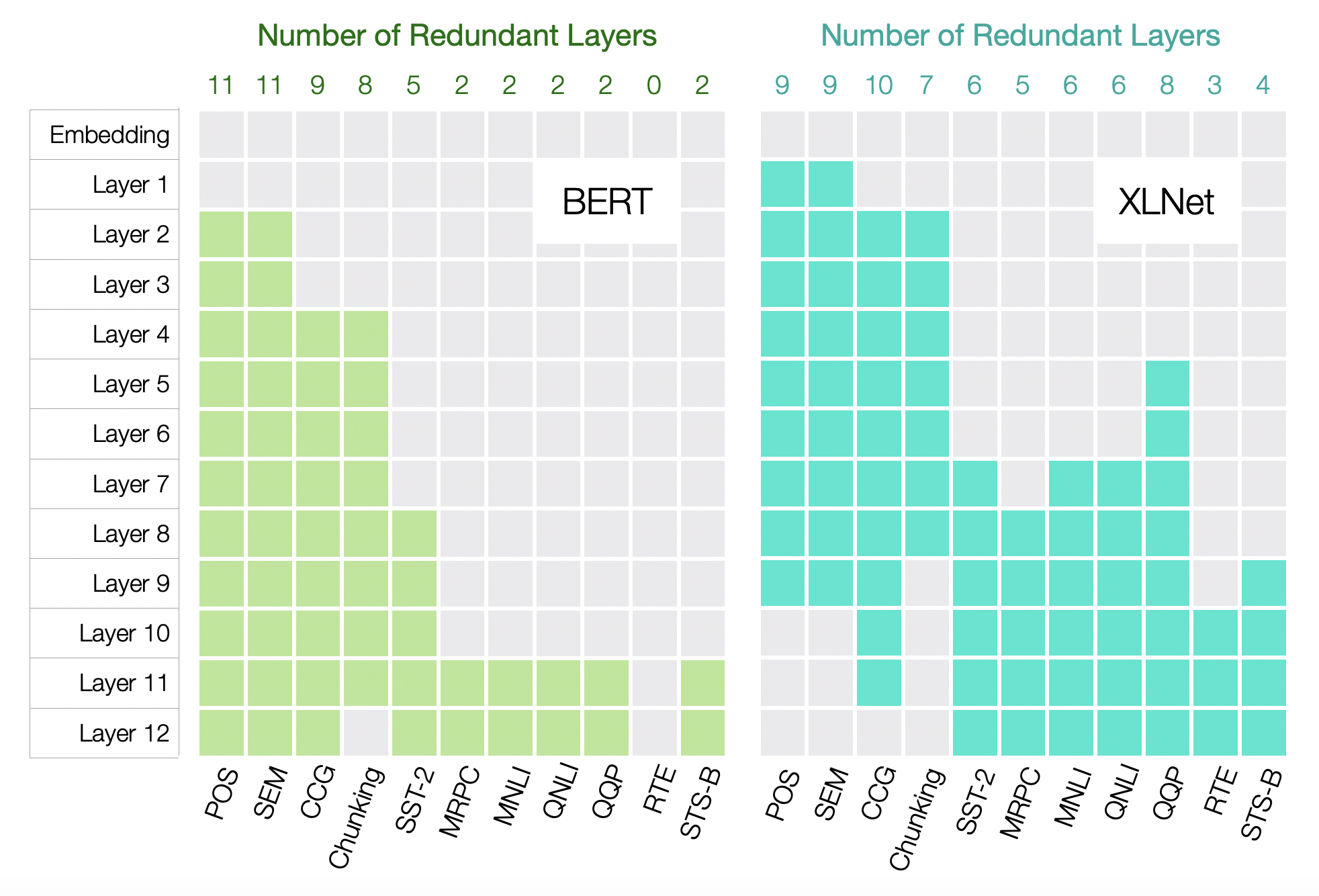}
		\caption{Task-specific layer-wise redundant layers represented by the colored blocks. Appendix \ref{subsec:appendix-task-specific-layer} presents fine-grained graphs for a few tasks.}
		\label{fig:summary-layerwise}
	\end{figure}
	Figure \ref{fig:summary-layerwise} shows the redundant layers that perform within a 1\% performance threshold with respect to the oracle on each task. We found high layer-level redundancy for sequence labeling tasks. There are up to 11 redundant layers in BERT and up to 10 redundant layers in XLNet, across different tasks.
	%\nd{Didn't we say in the intro that XLNet is more redundant?} 
	% HS: taht was for neuron-level which is a different case
	% FD: also here we are talking only about core linguistic tasks
	This is expected, because the sequence labeling tasks considered here 
	%\hll{under study} 
	are core language tasks, and the information related to %\sout{core linguistic phenomenon} 
	them 
	is spread across the network. Comparing models, we found such core 
	%\sout{linguistic} 
	language information to be distributed amongst fewer layers in XLNet.
	
	\paragraph{Substantially less amount of redundancy for higher-level tasks}
	The amount of redundancy is substantially lower for sequence classification tasks, with \emph{RTE} having the least number of redundant layers in both models. Especially in BERT, we did not find any layer that matched the oracle performance 
	%\hll{showing that information important to solve the RTE task is spread across different layers.} % this is true but not relevant in the context 
	for RTE.  
	%redundant to perform the RTE task well. 
	It is interesting to observe that all the sequence classification tasks are learned at higher layers and none of the lower layers were found to be redundant. 
	%Overall, t
	These results are 
	%\sout{understandable} 
	intuitive given that 
	%these 
	the sequence classification
	tasks require complex linguistic 
	%\sout{information} 
	knowledge, such as long range contextual dependencies, 
	%\sout{which is present only in} 
	which are only learned at the higher-layers of the model. Lower layers do not have the sufficient sentence-level context to perform these tasks well.
	
	%\paragraph{XLNet versus BERT}
	\paragraph{XLNet is more redundant than BERT}
	%\nd{there was an XLNET versus BERT comparison 1 paragraph ago also. Either move that here or don't make paragraph heading}
	%\paragraph{Discussion}
	%\hs{would discussion be better?}
	While XLNet has slightly fewer redundant layers for sequence labeling tasks, on average across all downstream tasks it shows high layer-level task-specific redundancy. %We had similar observation in analyzing general neuron-level redundancy.
	%In line with our findings about general neuron-level redundancy, XLNet is more redundant than BERT in layer-wise task-specific redundancy (see the higher number of redundant layers of XLNet compared to BERT for sequence classification tasks). 
	Having high redundancy for sequence-level tasks reflects that 
	%This may also mean that 
	XLNet learns the higher-level concepts much earlier in the network and this information is then passed to all the subsequent layers. 
	This 
	also showcases that XLNet is a much better candidate for model compression where several higher layers can be pruned with marginal loss in performance, as shown  
	%and size reduction when transfer learning as shown 
	by 
	\newcite{Sajjad2020PoorMB}. %\nd{This is a discussion item and being mixed within main result - do some moving around in this section so that it appears in the end}
	
	% The difference in redundancy patterns of XLNet and BERT may be arising from the differences in the overall design and training regime of the two 
	% %language 
	% models. Another 
	% %interesting % i am not sure if its interesting
	% observation 
	% %\sout{here} 
	% is that the higher layers of XLNet are less useful for core linguistic tasks. It could be that the higher layers in XLNet are better optimized for higher-level concepts and thus they are less suitable to perform core linguistic tasks.
	
	\subsection{Neuron-level Redundancy}
	\label{subsec:neuron-task-specific-redundancy}
	
	%\hll{Here, we now extend our analysis and explore whether neurons preserve task-specific information redundantly. Unlike layers, it is combinatorially intractable to exhaustively try all possible neuron permutations that can carry out a downstream task. We therefore aim at extracting only one minimal set of neurons that suffice the purpose, and consider the remaining neurons redundant or irrelevant for the task at hand.}
	
	Pretrained models being a universal feature extractor contain redundant information with respect to a downstream task. 
	%\hll{We now extend our analysis and explore whether neurons preserve task-specific information redundantly}
	We hypothesize that they may also contain information that is not necessary for the underlying task. In task-specific neuron analysis, we consider both redundant and irrelevant neurons as redundancy with respect to a task.
	Unlike layers, it is combinatorially intractable to exhaustively try all possible neuron permutations that can carry out a downstream task. We therefore aim at extracting only one 
	%set of 
	minimal set of neurons that suffice the purpose, and consider the remaining neurons redundant or irrelevant for the task at hand.
	%\yb{I'm again confused by the distinction between redundant and irrelevant. I think I get it, but not sure. }  
	
	Formally, given a task and a set of neurons from a model, we perform feature selection to identify a minimal set of neurons that match the oracle performance. To accomplish this,
	%
	%\sout{We model the task of identifying neuron-level redundancy as a feature selection problem.  Given a task and a set of neurons from a model, we perform feature selection to identify a minimal set of neurons that match the oracle performance. 
	%want to 
	%identify a minimum set of neurons which are sufficient for solving the given task. The %rest of the 
	%remaining neurons are either redundant \sout{with} \hll{to} the selected neurons or irrelevant for the task.
	%\yb{ah, yes, now I get it. maybe the problem was just with me.}
	%
	%
	we use the Linguistic Correlation Analysis method~\cite{dalvi:2019:AAAI} to 
	ranks neurons with respect to a downstream task, 
	%analyze task-specific neuron-level redundancy, 
	referred as \texttt{FS} (feature selector) henceforth. %\texttt{FS} 
	For each downstream task, we concatenate representations from all layers $L$ and use \texttt{FS} to extract a minimal set of top ranked neurons that maintain the oracle performance, within a defined threshold. %on the downstream task. 
	Oracle is the task-specific classification performance obtained using all the neurons for training. The minimum set allows us to answer how many neurons are redundant and irrelevant to the given task. %\nd{But this does not tell how many neurons are redundant}
	Tables \ref{tab:sequence-labeling-neurons} and \ref{tab:sequence-classification-neurons} show the minimum set of top neurons for each task that maintains at least 97\% of the oracle performance.
	%\nd{We had a threshold of 1\% before - why 3\% here?}
	
	\paragraph{Complex core language tasks require more neurons}
	CCG and Chunking are relatively complex tasks compared to POS and SEM. 
	On average across both models, these complex tasks 
	%y CCG and Chunking 
	require more neurons than POS and SEM. It is interesting to see that the size of minimum neurons set is correlated with the complexity of the task.
	%requires 
	%the least 
	%less number of neurons. %\nd{in the other paper with a threshold of 0\% CCG requires more neurons than Chunking also here XLNet requires more.}
	%Comparing various sequence labeling tasks, For sequence classification tasks, depending on the complexity of the sequence labeling task, we found  
	
	\begin{table}[t]
		\footnotesize
		\begin{subtable}{0.48\linewidth}
			\centering
			\resizebox{\linewidth}{!}{
				\begin{tabular}{l|c}
					\toprule
					Task & \# Neurons \\
					\midrule
					POS & 290 \\
					SEM & 330 \\
					CCG & 330 \\
					Chunk. & 750 \\
					\bottomrule
				\end{tabular}
			}
			\caption{BERT}
		\end{subtable}%
		\begin{subtable}{0.48\linewidth}
			\centering
			\resizebox{\linewidth}{!}{
				\begin{tabular}{l|c}
					\toprule
					Task & \# Neurons \\
					\midrule
					POS & 280 \\
					SEM & 290 \\
					CCG & 690 \\
					Chunk. & 660 \\
					\bottomrule
				\end{tabular}
			}
			\caption{XLNet}
		\end{subtable}
		\caption{Task-specific neuron-level analysis for sequence labeling tasks.}
		
		\label{tab:sequence-labeling-neurons}
	\end{table}
	
	\begin{table}[t]
		\footnotesize
		\begin{subtable}{.48\linewidth}
			\centering
			\resizebox{\linewidth}{!}{
				\begin{tabular}{l|c}
					\toprule
					Task & \# Neurons \\
					\midrule
					SST-2 & 30 \\
					MRPC & 190 \\
					MNLI & 30 \\
					QNLI & 40 \\
					QQP & 10 \\
					RTE & 320 \\
					STS-B & 290 \\
					\bottomrule
				\end{tabular}
			}
			\caption{BERT}
		\end{subtable}%
		\begin{subtable}{.48\linewidth}
			\centering
			\resizebox{\linewidth}{!}{
				\begin{tabular}{l|c}
					\toprule
					Task & \# Neurons \\
					\midrule
					SST-2 & 70 \\
					MRPC & 170 \\
					MNLI & 90 \\
					QNLI & 20 \\
					QQP & 20 \\
					RTE & 400 \\
					STS-B & 300 \\
					\bottomrule
				\end{tabular}
			}
			\caption{XLNet}
		\end{subtable} 
		\caption{Task-specific neuron-level analysis for sequence classification tasks.}
		\label{tab:sequence-classification-neurons}
	\end{table}

	\paragraph{Less task-specific redundancy for core linguistic tasks compared to higher-level tasks}
	While the minimum set of neurons per task consist of a small percentage of total neurons in the network, the core linguistic 
	%\sout{(sequence labeling)}
	%\nd{you are implying that all sequence labeling tasks are core lingustic tasks} 
	tasks require substantially more neurons compared to higher-level 
	%(sequence classification) 
	tasks (comparing Tables \ref{tab:sequence-labeling-neurons} and \ref{tab:sequence-classification-neurons}). It is remarkable 
	%\sout{to find out} 
	that some sequence-level tasks require as few as only 10 neurons to obtain desired performance. One reason for the large difference in the size of minimum set of neurons
	%between these two classes of tasks 
	%in sequence labeling tasks and seq
	%for this may be 
	could be the nature of tasks, since core linguistic tasks are word-level tasks, a much higher capacity is required in the pretrained model to store the knowledge for all of the words. While in the case of sequence classification tasks, the network learns to filter and mold the features to form fewer ``high-level'' sentence features. 
	%
	%\hll{This hypothesis is further supported from Figure \ref{fig:summary-layerwise} where only higher layers were able to achieve a performance close to oracle. In other words, the sequence-level tasks are learned at higher layers where higher-level phenomena are learned and captured in fewer neurons.}
	%which means all of the sequence classification tasks have 
	%
	%We see very similar results at neuron-level for both BERT and XLNet across both task sets. The average redundancy is much lower in the case of sequence labeling tasks (higher number of neurons selected) as compared to sequence classification tasks. 
	%
	%in a cone-fashion; 
	%An interesting direction could be pruning the models in a cone-fashion; 
	%apply aggressive pruning on 
	%remove significantly more features in the 
	%higher layers than the lower layers might allow maintaining a high capacity for word level information in the network, without wasting a lot of capacity in the higher layers where a few, rich features are sufficient for a given task.

	%\section{Application: Efficient Transfer Learning}
	\section{Efficient Transfer Learning}
	\label{sec:transfer-learning}
	%One direct use case of the redundancy analysis presented in the previous sections is removing redundant parts of a pre-trained language model to enable more efficient transfer learning, both in terms of computation and memory. 
	%The redundancy in pre-trained models can be exploited for efficient transfer learning. 
	In this section, we build upon the redundancy analysis presented in the previous sections and propose a novel method for efficient feature-based transfer learning. 
	In a typical feature-based transfer learning setup, contextualized embeddings are first extracted 
	%(from a specific layer or all layers) 
	from a pretrained model, and then a classifier is trained on the embeddings towards the downstream NLP task. The bulk of the computational expense is incurred from the following sources: 
	\begin{itemize}[leftmargin=*]
		\setlength\itemsep{-0.03em}
		\item A full forward pass over the pretrained model to extract the contextualized vector,
		%which is
		a costly affair given the large number of parameters.
		\item Classifiers with large contextualized vectors are: a) cumbersome to train, b) inefficient during inference, and c) may be sub-optimal when supervised data is insufficient \cite{hameed}.
	\end{itemize}
	%
	%i) a full forward pass over the pre-trained model to extract the contextualized vector, which is a costly affair given the large number of model parameters, ii)  classifiers with large contextualized vectors are: a) cumbersome to train, b) inefficient during inference, and c) may be sub-optimal when supervised data is insufficient \cite{hameed}.
	%\begin{itemize}
	%	\item it requires a full forward pass over the pre-trained model to extract the contextualized vectors, which is a costly affair given the large number of model parameters. For example, a forward pass in BERT large requires computing 340 million parameters.
	%	\item classifiers with large contextualized vectors are a) cumbersome to train, b) inefficient during inference, and c) may be sub-optimal when supervised data is insufficient \cite{hameed}.
	%\end{itemize}
	%
	%We propose a novel efficient transfer learning method based on the presented analysis that targets %both of 
	%these inefficiencies by removing both general and task-specific redundant parts of the pretrained language model. %Our proposed method is 
	%
	\noindent We propose a three step process to target these two sources of computation bottlenecks: 
	\begin{enumerate}[leftmargin=*]
		\setlength\itemsep{-0.03em}
		\item Use the task-specific layer-classifier (%as presented in 
		Section~\ref{subsec:layer-task-specific-redundancy}) to select the lowest layer that maintains oracle performance. Differently from the analysis, a concatenation of all layers until the selected layer is used instead of just the individual layers.
		\item Given the contextualized embeddings extracted in the previous step, use \texttt{CC} (%as defined in
		Section \ref{subsec:neuron-general-redundancy}) to filter-out redundant neurons. %and reduce the size of the contextualized embeddings
		\item Apply \texttt{FS} (Section~\ref{subsec:neuron-task-specific-redundancy}) to select a minimal set of neurons that are needed to achieve optimum performance on the task.
	\end{enumerate}
	
	The three steps explicitly target task-specific layer redundancy, general neuron redundancy and task-specific neuron redundancy respectively. 
	We refer to Step 1 as \texttt{LayerSelector} (\texttt{LS}) and Step 2 and 3 as \texttt{CCFS} (Correlation clustering + Feature selection) later on.
	%In the following sections, we present the results on the sequence labeling and sequence classification tasks. %\hll{It is worth mentioning here that the trade-off between loss in accuracy and efficiency can be controlled through this threshold which can be adjusted to serve faster turn-around or better performance.}
	For all experiments, we use a performance threshold of 1\% for \texttt{LS} and \texttt{CCFS} each. It is worth mentioning that the trade-off between loss in accuracy and efficiency can be controlled through these thresholds, which can be adjusted to serve faster turn-around or better performance.

	\begin{table}[t]
		\centering
		\resizebox{\columnwidth}{!}{
			\begin{tabular}{l|cc|cc}
				\toprule
				& \multicolumn{2}{c}{Sequence} & \multicolumn{2}{c}{Sequence} \\
				& \multicolumn{2}{c}{Classification} & \multicolumn{2}{c}{Labeling} \\
				\cmidrule{1-5}
				& BERT    &   XLNet &  BERT  &  XLNet  \\
				\cmidrule{1-5}
				Oracle         & 93.0\%  &  93.4\% & 85.5\% &  84.8\% \\ 
				Neurons        &      \multicolumn{4}{c}{9984}        \\
				\cmidrule{2-5}
				\texttt{LS}    & 92.3\%  &  93.2\% & 85.0\% &  84.5\% \\ 
				Layers         &  5.3   &    2.5  &  11.6  &  8.1    \\
				\cmidrule{2-5}
				\texttt{CCFS}  & 92.0\%  &  92.2\% & 84.0\% & 84.0\%  \\
				Neurons        & 425     &  400    & 90     &  150    \\
				\cmidrule{2-5}
				\% Reduct.     & 95.7\%$\downarrow$ & 96.0\%$\downarrow$ & 99.0\%$\downarrow$ & 98.5\%$\downarrow$ \\
				\bottomrule
			\end{tabular}
		}
		\caption{Average results 
			%of sequence labeling and sequence classification tasks 
			using \texttt{LS} and \texttt{CCFS} with performance thresholds of 1\% for each. Oracle is using a concatenation of all layers. \emph{Layers} shows the average number of selected layers. \textit{Neurons} are the final number of neurons (features) used for classification. \emph{$\%$ Reduct.} shows the percentage reduction in neurons compared to the full network.}
		\label{tab:results_lsccfs}
	\end{table}
	
	\subsection{Results}
	Table \ref{tab:results_lsccfs} presents the average results on all sequence labeling and sequence classification tasks. Detailed per-task results are provided in Appendix \ref{subsec:appendix-transfer-learning}. As expected from our analysis, a significant portion of the network can be pruned by \texttt{LS} for sequence labeling tasks, using less than 6 layers out of 13 (Embedding + 12 layers) for BERT and less than 3 layers for XLNet. Specifically, this reduces the parameters required for a forward pass for BERT by 65\% for POS and SEM, and 33\% for CCG and 39\% for Chunking. On XLNet, \texttt{LS} led to even larger reduction in
	%pruning, with the number of 
	parameters; 
	%reduced by 
	70\% for POS and SEM, and 65\% for CCG and Chunking. The results were less pronounced for sequence classification tasks, 
	%with \texttt{LS} using 11.6 layers on average out of 13 for BERT, and XLNet using 8.1 layers. 
	with \texttt{LS} using 11.6 layers for BERT and 8.1 layers for XLNet on average, out of 13 layers. 
	
	Applying \texttt{CCFS} on top of the reduced layers led to another round of significant efficiency improvements. 
	The number of neurons needed for the final classifier reducing to just 5\% for sequence labeling tasks and 1.5\% for sequence classification tasks. The final number of neurons is surprising low for some tasks compared to the initial $9984$, with some tasks like QNLI using just 10 neurons.
	
	More concretely, taking the POS task as an example: the pre-trained oracle BERT model has 9984 features and 110M parameters. \texttt{LS} reduced the feature set to 2304 (embedding + 2 layers) and the number of parameters used in the forward pass to 37M. \texttt{CCFS} further reduced the feature set to 300, maintaining a performance close to oracle BERT's performance on this task (95.2\% vs.\ 93.9\%).
	
	%\nd{This paragraph is nice except that this section is about efficiency and not analysis. I wonder if we should move analysis to previous section and keep this section short?}
	% i tired it before and moved it back. we are not talking about fewer layers from the classification point of view.
	An interesting observation in Table \ref{tab:results_lsccfs} is that the sequence labeling tasks require \textit{fewer layers} but \textit{a higher number of features}, while sequence classification tasks follow the opposite pattern. As we go deeper in the network, the neurons are much more richer and tuned for the task at hand, and only a few of them are required compared to the much more word-focused neurons in the lower layers. These observations suggest pyramid-shaped architectures that have wider lower layers and narrow higher layers.
	%may perform well for NLP applications. 
	Such a design choice 
	%This 
	leads to significant savings of capacity in higher layers where a few, rich neurons are sufficient for good performance. In terms of neuron-based compression methods,
	%to prune models, 
	these findings propose aggressive pruning of higher layers while preserving the lower layers in building smaller and 
	accurate compressed models.

	\begin{figure}[t]
		\centering
		\includegraphics[width=0.8\linewidth]{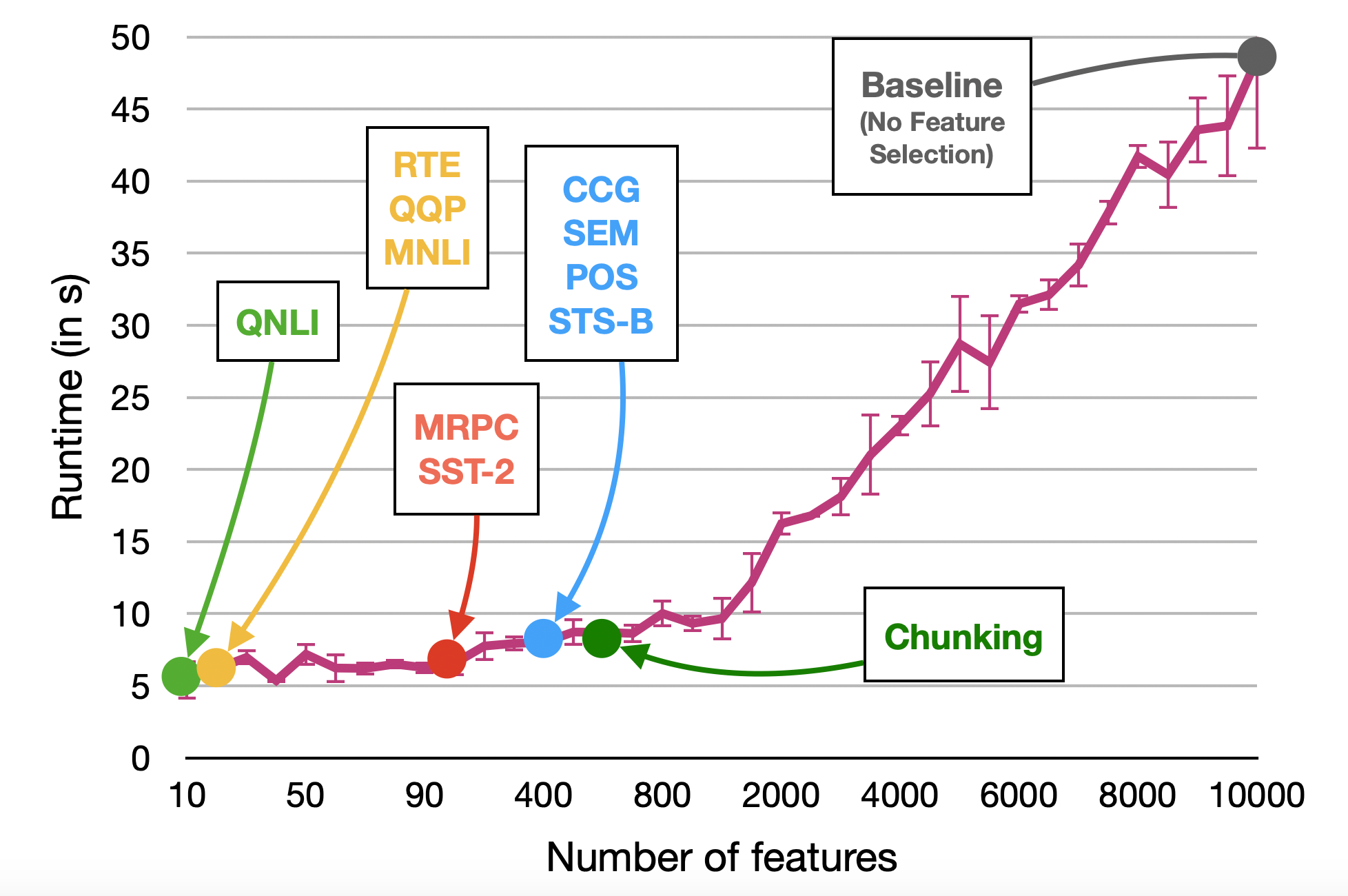}
		\caption{BERT: Runtime of the classifier w.r.t. number of neurons (features). 
			%for 100K instances as we increase the number of features (neurons) used as input to the classifier. 
			The dots 
			on the line 
			mark the number of features selected by our method. 
			%for each of the tasks in BERT. We can see that for all of the tasks, we use less than one-fifth of the time taken with all the features. 
			Note that the X-axis is not linear, the lower half of the spectrum has been stretched for clarity.}
		\label{fig:classifier_timing}
	\end{figure}
	
	\subsection{Efficiency Analysis}
	While the algorithm boosts the theoretical efficiency in terms of the number of parameters reduced and the final number of features, it is important to analyze how this translates to real world performance. Using \texttt{LS} leads to an average speed up of $2.8{\tt x}$ and $6.2{\tt x}$ with BERT and XLNet respectively on sequence labeling tasks. On sequence classification tasks, the average speed ups are $1.1{\tt x}$ and $1.6{\tt x}$ with BERT and XLNet respectively. Detailed %per-task 
	%speed ups 
	results
	are provided in Appendix \ref{subsec:appendix-pretrained-timing}.
	
	For the classifier built on the reduced set, we simulate a test scenario with 100,000 tokens and compute the total runtime for 10 iterations of training. The numbers were computed on a 6-core 2.8 GHz AMD Opteron Processor 4184, and were averaged across 3 runs. Figure \ref{fig:classifier_timing} shows the runtime of each run (in seconds) against the number of features selected. %We can see that for the tasks in BERT, 
	%We can see t
	The runtime of the classifier reduced from 50 to 10 seconds in the case of BERT. %the classifier finished running in less than 10 seconds compared to the original runtime of around 50 seconds. 
	The $5{\tt x}$ speedup can be very useful in a heavy-use scenarios where the classifier is queried a large number times in a short duration.
	
	\paragraph{Training time efficiency: } Although the focus of the current application is to improve inference-time efficiency, it is nevertheless important to understand how much computation complexity is added during training time. Let $T$ be the total number of tokens in our training set, and $N$ be the total number of neurons across all layers in a pre-trained model. The application presented in this section consists of 5 steps. 
	
	\begin{enumerate}
		\item Feature extraction from pre-trained model: Extraction time scales linearly with the number of tokens $T$.
		\item Training a classifier for every layer \texttt{LS}: With a constant number of neurons $N$, training time per layer scales linearly with the number of tokens $T$.
		\item Correlation clustering \texttt{CC}: With a constant number of neurons $N$, running correlation clustering scales linearly with the number of tokens $T$.
		\item Feature ranking: This step involves training a classifier with the reduced set of features, which scales linearly with the number of tokens $T$. Once the classifier is trained, the weights of the classifier are used to extract a feature ranking, with the number of weights scaling linearly with the number of selection neurons $N$.
		\item Minimal feature set: Finding the minimal set of neurons is a brute-force search process, starting with a small number of neurons. For each set of neurons, a classifier is trained, the time for which scales linearly with the total number of tokens $T$. As the feature set size increases, the training time also goes up as described in Figure \ref{fig:classifier_timing}.
	\end{enumerate}
	
	Appendix \ref{subsec:appendix-training-time} provides additional experiments and results used to analyze the training time complexity of our application.

	\section{Conclusion and Future Directions}
	\label{sec:conclusion}
	
	%\nd{The introduction is like, we define the notion of redundancy but conclusion is like we present methods to study redundancy. }
	We defined a notion of redundancy and 
	%in pretrained models and introduced several methods to analyze 
	%different forms of redundancy in various parts of pretrained models. More specifically, we 
	analyzed pre-trained models for 
	general redundancy and task-specific redundancy exhibited at layer-level and at individual neuron-level. 
	Our analysis on general redundancy showed that i) adjacent layers are most redundant in the network with an exception of final layers which are close to the objective function,
	%softmax layer,
	and ii) up to 85\% and 92\% neurons are redundant in BERT and XLNet respectively. We further showed that networks exhibit varying amount of task-specific redundancy; higher layer-level redundancy for core language tasks compared to sequence-level tasks. We found that at least 92\% of the neurons are redundant with respect to a downstream task.
	Based on our analysis, we proposed an efficient transfer learning procedure that directly targets layer-level and neuron-level redundancy to achieve efficiency in feature-based transfer learning.
	
	While our analysis is helpful in understanding pretrained models, it suggests interesting research directions towards building compact models and models with better architectural choices. For example, a high amount of neuron-level redundancy in the same layer suggests that layer-size compression might be more effective in reducing the pretrained model size while preserving oracle performance. Similarly, our finding that core-linguistic tasks are learned at lower-layers and require a higher number of neurons, while sequence-level tasks are learned at higher-layers and require fewer neurons, suggests pyramid-style architectures that have wide lower layers and compact higher layers and may result in smaller models with performance competitive with large models. %with less number of parameters. reduce the size of the pretrained models.
	
	\section*{Acknowledgements}
	This research was carried out in collaboration between the HBKU Qatar Computing Research Institute (QCRI) and the MIT Computer Science and Artificial Intelligence Laboratory (CSAIL). Y.B.\  was also supported by the Harvard Mind, Brain, and Behavior Initiative (MBB).

	\bibliography{emnlp2020,acl2020}
	\bibliographystyle{acl_natbib}
	
	\newpage
	\appendix
	\section{Appendices}
	\subsection{Data}
	\label{subsec:appendix-data}
	For Sequence labeling tasks, we use the first 150,000 tokens for training, and standard development and test data for all of the four tasks (POS, SEM, CCG super tagging and Chunking). The links to all datasets is provided in the code README instructions. The statistics for the datasets are provided in Table \ref{tab:word_data_statistics}.
	
	\begin{table}[ht]									
		\centering					
		\begin{tabular}{l|cccc}									
			\toprule									
			Task    & Train & Dev & Test & Tags \\		
			\midrule
			POS & 149973 & 44320 & 47344 & 44 \\
			SEM & 149986 & 112537 & 226426 & 73 \\
			Chunking &  150000 &  44346 &  47372 & 22 \\
			CCG &  149990 & 45396 & 55353 & 1272 \\
			\bottomrule
		\end{tabular}
		\caption{Data statistics (number of tokens) on training, development and test sets used in the experiments and the number of tags to be predicted}
		\label{tab:word_data_statistics}						
	\end{table}
	
	For the sequence classification tasks, we study tasks from the GLUE benchmark~\citep{wang-etal-2018-glue}, namely i) sentiment analysis (SST-2) using the Stanford sentiment treebank \cite{socher-etal-2013-recursive}, ii) semantic equivalence classification using the Microsoft Research paraphrase corpus (MRPC) \cite{dolan-brockett-2005-automatically}, iii) natural language inference 
	%using the MultiNLI 
	corpus (MNLI) \cite{williams-etal-2018-broad}, iv) question-answering NLI (QNLI) using the SQUAD dataset \cite{rajpurkar-etal-2016-squad}, iv) question pair similarity using the Quora Question Pairs\footnote{\url{http://data.quora.com/First-Quora-Dataset-Release-Question-Pairs}} dataset (QQP), v) textual entailment using recognizing textual entailment dataset(RTE) \cite{Bentivogli09thefifth}, and vi) semantic textual similarity using the STS-B dataset \cite{cer-etal-2017-semeval}. The statistics for the datasets are provided in Table \ref{tab:sentence_data_statistics}.
	
	\begin{table}[ht]									
		\centering					
		\begin{tabular}{l|cc}									
			\toprule									
			Task    & Train & Dev  \\		
			\midrule
			SST-2 &  67349 &   872  \\
			MRPC  &   3668 &   408  \\
			MNLI  & 392702 &  9815  \\
			QNLI  & 104743 &  5463  \\
			QQP   & 363846 & 40430  \\
			RTE   &   2490 &   277  \\
			STS-B &   5749 &  1500 \\
			\bottomrule
		\end{tabular}
		\caption{Data statistics (number of sequences) on the official training and development sets used in the experiments. All tasks are binary classification tasks, except for STS-B which is a regression task. Recall that the test sets are not publicly available, and hence we use 10\% of the official train as development, and the official development set as our test set. Exact split information is provided in the code README.}
		\label{tab:sentence_data_statistics}						
	\end{table}

	\subsection{General Neuron-level Redundancy}
	\label{subsec:appendix-general-neuron}
	Table \ref{tab:appendix-cc-alltasks} presents the detailed results for the illustration in Figures  \ref{subfig:bert-cc-reduction} and \ref{subfig:xlnet-cc-reduction}. As a concrete example, 6 out of 12 tasks (POS, SEM, CCG, Chunking, SST-2, STS-B) can do away with more than $85\%$ reduction in the number of neurons (threshold=0.7) with very little loss in performance. 

\begin{figure*}[t]
		\centering
		\includegraphics[width=\linewidth]{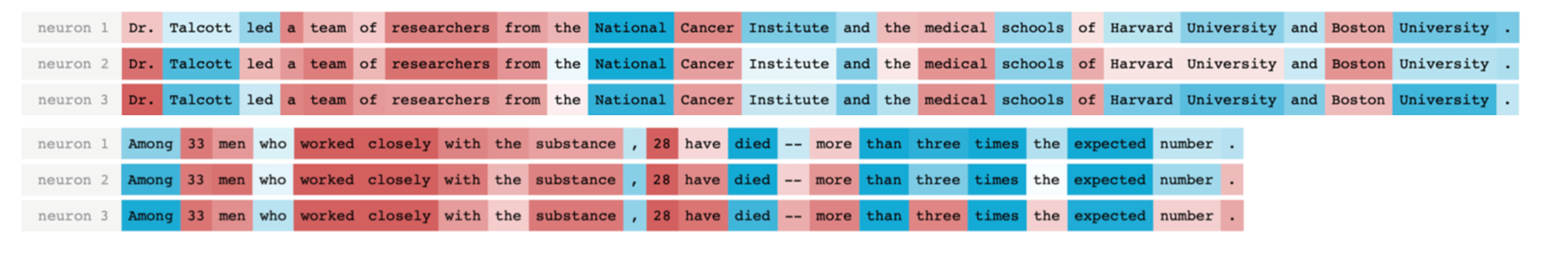}
		\caption{Redundant neurons as clustered by \textit{correlation clustering} on two sentences. The dark red and dark blue refer to high negative and positive activation values respectively.}
		\label{fig:sentence_heatmap}
	\end{figure*}

	Figure \ref{fig:sentence_heatmap} visualizes heatmaps of a few neurons that belong to the same cluster built using \texttt{CC} at $c_t=0.3$ as a qualitative example of a cluster.
	
	\begin{table*}[t]
		\begin{subtable}{0.99\linewidth}
			\centering
			\resizebox{\linewidth}{!}{
				\begin{tabular}{r|cccc|ccccccc|c}
					\toprule
					Threshold & POS & SEM & CCG & Chunking & SST-2 & MRPC & MNLI & QNLI & QQP & RTE & STS-B & Average \\
					\midrule
					0.0 & 9984 & 9984 & 9984 & 9984 & 9984 & 9984 & 9984 & 9984 & 9984 & 9984 & 9984 & 9984 \\
					0.0 & 95.7\% & 92.0\% & 89.8\% & 94.5\% & 90.5\% & 85.8\% & 81.7\% & 90.3\% & 91.2\% & 70.0\% & 89.5\% & 88.3\% \\
					0.1 & 6841 & 6809 & 6844 & 6749 & 7415 & 9441 & 9398 & 8525 & 8993 & 9647 & 8129 & 8072 \\
					0.1 & 95.4\% & 92.3\% & 90.3\% & 94.8\% & 89.8\% & 86.3\% & 81.7\% & 90.2\% & 91.2\% & 69.3\% & 89.7\% & 88.3\% \\
					0.2 & 4044 & 4045 & 4052 & 4008 & 6207 & 8486 & 8376 & 7225 & 7697 & 8705 & 6377 & 6293 \\
					0.2 & 95.9\% & 92.9\% & 90.6\% & 95.0\% & 90.6\% & 86.8\% & 81.7\% & 90.1\% & 91.2\% & 69.0\% & 89.6\% & 88.5\% \\
					0.3 & 2556 & 2566 & 2570 & 2573 & 4994 & 7328 & 7049 & 6131 & 6413 & 7157 & 4949 & 4935 \\
					0.3 & 96.2\% & 93.1\% & 91.3\% & 95.1\% & 90.6\% & 86.0\% & 81.8\% & 89.9\% & 91.1\% & 67.1\% & 89.5\% & 88.3\% \\
					0.4 & 1729 & 1752 & 1729 & 1709 & 3812 & 5779 & 5681 & 4961 & 5077 & 5587 & 3674 & 3772 \\
					0.4 & 96.2\% & 93.3\% & 91.4\% & 95.2\% & 90.4\% & 86.5\% & 81.7\% & 89.4\% & 91.0\% & 67.5\% & 89.3\% & 88.4\% \\
					0.5 & 1215 & 1190 & 1221 & 1217 & 2746 & 4420 & 4289 & 3747 & 3789 & 4241 & 2721 & 2800 \\
					0.5 & 96.4\% & 93.2\% & 91.6\% & 94.9\% & 90.3\% & 86.3\% & 81.6\% & 89.6\% & 91.1\% & 66.4\% & 89.0\% & 88.2\% \\
					0.6 & 876 & 869 & 873 & 876 & 1962 & 3287 & 3041 & 2712 & 2767 & 3170 & 1962 & 2036 \\
					0.6 & 96.2\% & 93.3\% & 91.5\% & 94.4\% & 90.0\% & 85.5\% & 81.8\% & 89.7\% & 91.1\% & 66.8\% & 88.8\% & 88.1\% \\
					0.7 & 792 & 789 & 792 & 795 & 1404 & 2258 & 2025 & 1867 & 1907 & 2315 & 1419 & 1488 \\
					0.7 & 96.2\% & 93.2\% & 91.6\% & 94.1\% & 89.8\% & 86.3\% & 81.7\% & 89.3\% & 91.1\% & 69.0\% & 87.8\% & 88.2\% \\
					0.8 & 764 & 758 & 762 & 748 & 982 & 1367 & 1239 & 1191 & 1226 & 1531 & 982 & 1050 \\
					0.8 & 96.1\% & 93.2\% & 91.3\% & 94.0\% & 89.2\% & 85.0\% & 80.6\% & 88.3\% & 90.0\% & 62.8\% & 82.6\% & 86.7\% \\
					0.9 & 443 & 378 & 429 & 357 & 778 & 812 & 798 & 797 & 814 & 854 & 785 & 659 \\
					0.9 & 95.6\% & 91.8\% & 89.9\% & 91.0\% & 56.5\% & 70.3\% & 53.2\% & 80.0\% & 77.6\% & 59.2\% & 32.5\% & 72.5\% \\
					\bottomrule
				\end{tabular}
			}
			\caption{BERT}
			\label{subtab:appendix-cc-alltasks-bert}
		\end{subtable} \\
		\\
		\begin{subtable}{0.99\linewidth}
			\centering
			\resizebox{\linewidth}{!}{
				\begin{tabular}{r|cccc|ccccccc|c}
					\toprule
					Threshold & POS & SEM & CCG & Chunking & SST-2 & MRPC & MNLI & QNLI & QQP & RTE & STS-B & Average \\
					\midrule
					0.0 & 9984 & 9984 & 9984 & 9984 & 9984 & 9984 & 9984 & 9984 & 9984 & 9984 & 9984 & 9984 \\
					0.0 & 96.2\% & 91.8\% & 90.6\% & 93.5\% & 93.2\% & 86.5\% & 78.9\% & 89.1\% & 87.4\% & 69.7\% & 89.0\% & 87.8\% \\
					0.1 & 9019 & 9021 & 9046 & 8941 & 7435 & 9206 & 7913 & 8056 & 5844 & 9931 & 9125 & 9006.75 \\
					0.1 & 96.3\% & 92.2\% & 90.7\% & 93.9\% & 93.0\% & 86.5\% & 80.3\% & 89.2\% & 89.7\% & 71.8\% & 89.0\% & 88.4\% \\
					0.2 & 5338 & 5392 & 5346 & 5302 & 6257 & 7685 & 6668 & 7393 & 4952 & 9244 & 8011 & 5344.5 \\
					0.2 & 96.2\% & 92.3\% & 90.5\% & 93.9\% & 93.0\% & 86.8\% & 80.4\% & 89.9\% & 90.2\% & 70.4\% & 88.9\% & 88.4\% \\
					0.3 & 3646 & 3651 & 3660 & 3606 & 5206 & 6241 & 5988 & 6613 & 4482 & 7635 & 6407 & 3640.75 \\
					0.3 & 96.2\% & 92.5\% & 91.0\% & 93.8\% & 92.9\% & 86.8\% & 80.8\% & 89.8\% & 90.1\% & 71.5\% & 88.7\% & 88.5\% \\
					0.4 & 2592 & 2571 & 2599 & 2573 & 4181 & 4896 & 5252 & 5583 & 3987 & 5996 & 4932 & 2583.75 \\
					0.4 & 96.3\% & 92.7\% & 90.8\% & 93.7\% & 93.1\% & 88.0\% & 81.0\% & 89.7\% & 90.1\% & 70.4\% & 88.5\% & 88.6\% \\
					0.5 & 1754 & 1746 & 1756 & 1758 & 3207 & 3675 & 4172 & 4426 & 3271 & 4573 & 3669 & 1753.5 \\
					0.5 & 96.5\% & 92.8\% & 91.3\% & 94.4\% & 93.2\% & 87.7\% & 80.8\% & 89.6\% & 90.1\% & 71.8\% & 88.3\% & 88.8\% \\
					0.6 & 1090 & 1085 & 1091 & 1072 & 2355 & 2549 & 2905 & 3248 & 2370 & 3346 & 2666 & 1084.5 \\
					0.6 & 96.7\% & 93.0\% & 91.8\% & 93.8\% & 93.1\% & 88.0\% & 81.0\% & 90.4\% & 90.0\% & 70.4\% & 88.4\% & 88.8\% \\
					0.7 & 833 & 833 & 830 & 824 & 1663 & 1735 & 1883 & 2224 & 1627 & 2348 & 1859 & 830 \\
					0.7 & 96.6\% & 93.0\% & 91.9\% & 93.2\% & 92.0\% & 88.2\% & 79.9\% & 90.1\% & 89.7\% & 71.1\% & 87.7\% & 88.5\% \\
					0.8 & 773 & 775 & 773 & 762 & 1127 & 1108 & 1189 & 1399 & 1091 & 1469 & 1232 & 770.75 \\
					0.8 & 96.5\% & 92.9\% & 91.9\% & 93.0\% & 92.4\% & 85.5\% & 77.3\% & 89.4\% & 87.4\% & 69.3\% & 84.5\% & 87.3\% \\
					0.9 & 470 & 412 & 471 & 414 & 799 & 790 & 805 & 839 & 791 & 832 & 801 & 441.75 \\
					0.9 & 96.0\% & 91.5\% & 91.0\% & 90.5\% & 84.4\% & 75.0\% & 65.8\% & 79.7\% & 88.3\% & 63.9\% & 46.6\% & 79.3\% \\
					\bottomrule
				\end{tabular}
			}
			\caption{XLNet}
			\label{subtab:appendix-cc-alltasks-xlnet}
		\end{subtable}
		\caption{Accuracies and number of neurons across all tasks after running \textit{correlation clustering}. Recall that the clustering is run without any task specific labels, and the evaluation is done across all tasks to analyze the efficacy of \textit{correlation clustering} as a method to remove redundant neurons.}
		\label{tab:appendix-cc-alltasks}
	\end{table*}
	
	\subsection{Task-Specific Layer-wise redundancy}
	\label{subsec:appendix-task-specific-layer}
	Tables \ref{tab:full_results_task_specific_layer_bert} and \ref{tab:full_results_task_specific_layer_bert} provide detailed results used to produce the illustrations in Figure \ref{fig:summary-layerwise}.
	
\begin{figure*}[t]
		\begin{subfigure}{.98\linewidth}
			\centering
			\includegraphics[width=\linewidth]{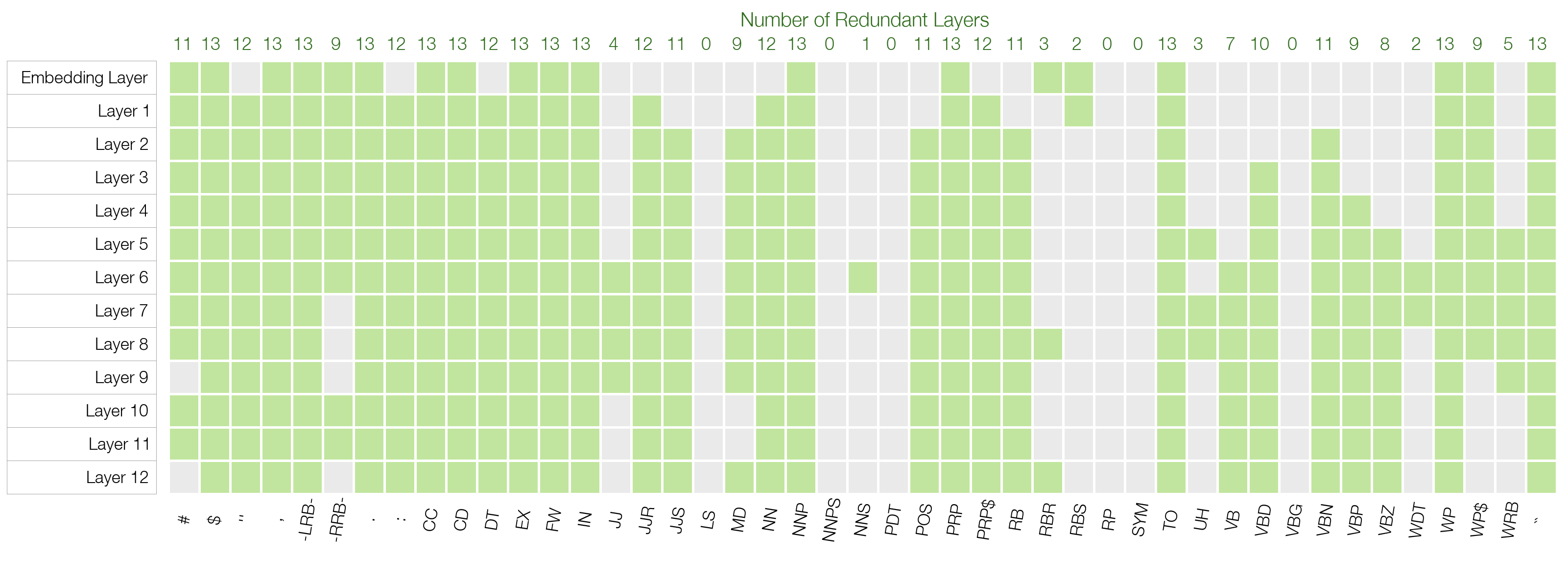}  
			\caption{BERT}
			\label{subfig:bert-layer-redundancy-pos}
		\end{subfigure}
		\begin{subfigure}{.98\linewidth}
			\centering
			\includegraphics[width=\linewidth]{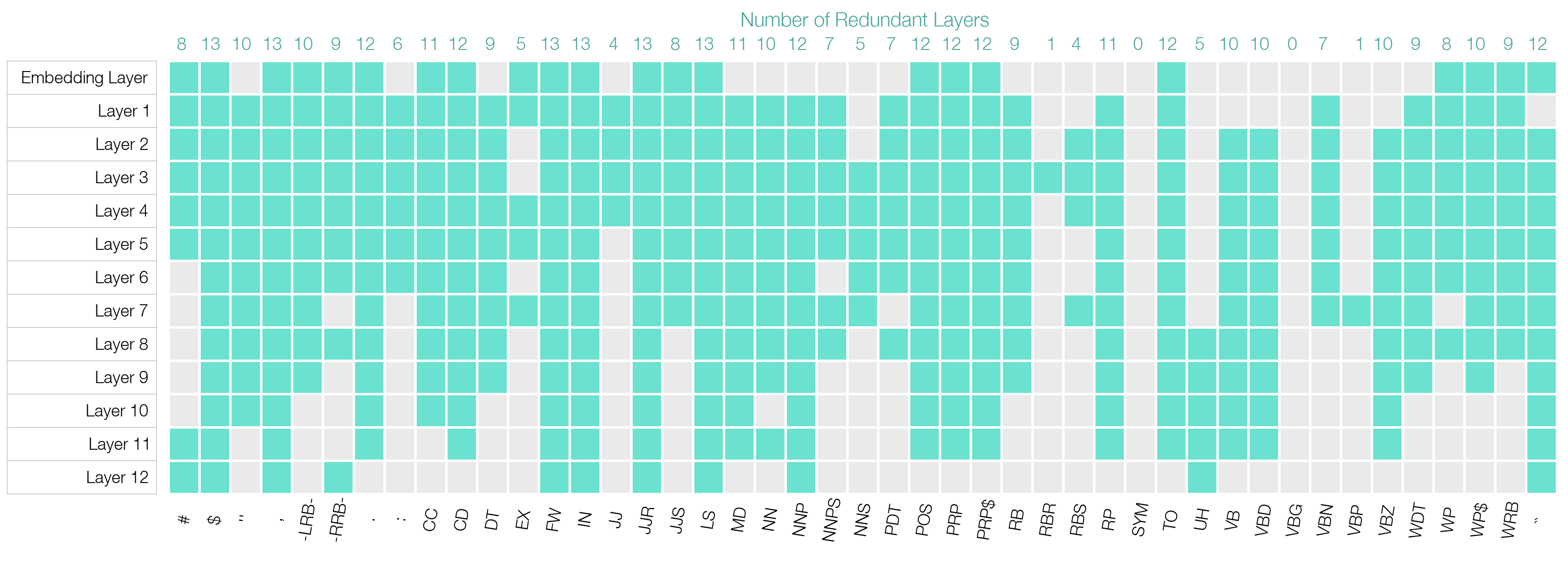}
			\caption{XLNet}
			\label{subfig:xlnet-layer-redundancy-pos}
		\end{subfigure}
		\caption{Layer-wise task specific redudancy for POS task. Redundant layers are represented by the colored blocks.}
		\label{fig:layer-redundancy-pos}
	\end{figure*}

\begin{figure*}[t]
	\begin{subfigure}{.98\linewidth}
		\centering
		\includegraphics[width=\linewidth]{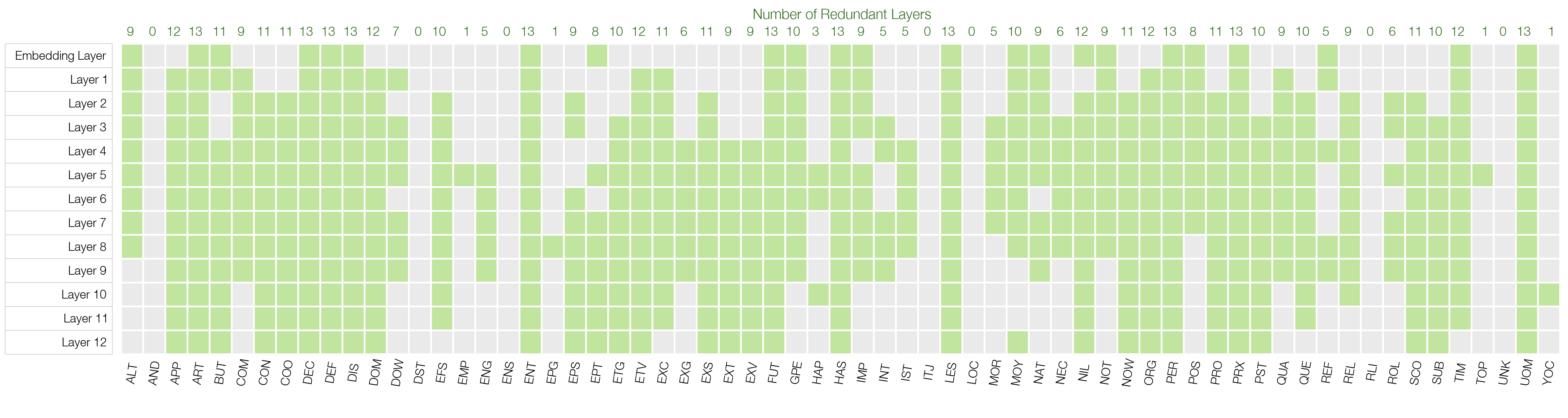}  
		\caption{BERT}
		\label{subfig:bert-layer-redundancy-sem}
	\end{subfigure}
	\begin{subfigure}{.98\linewidth}
		\centering
		\includegraphics[width=\linewidth]{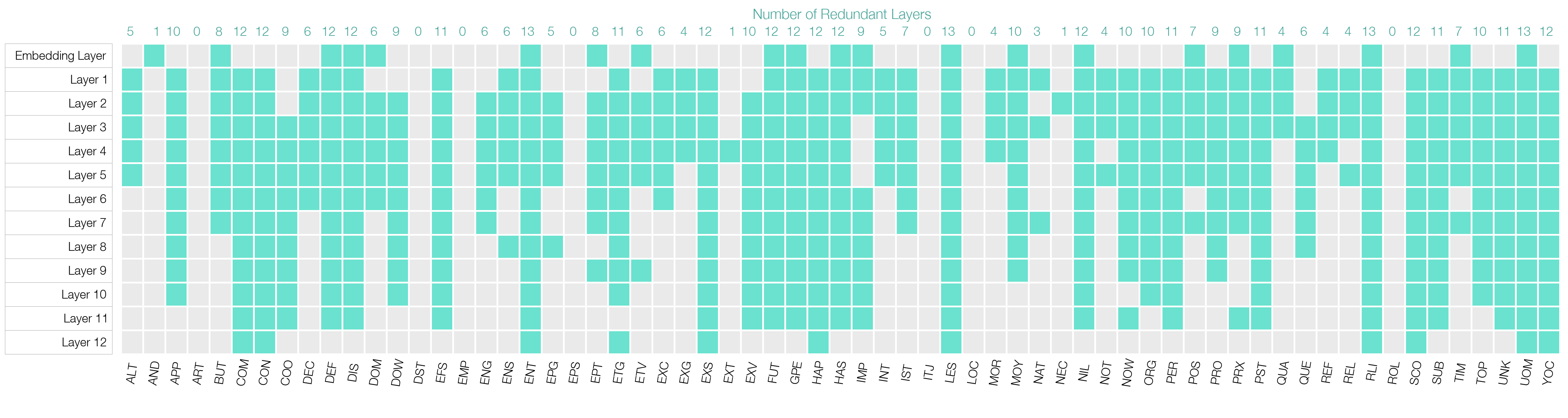}
		\caption{XLNet}
		\label{subfig:xlnet-layer-redundancy-sem}
	\end{subfigure}
	\caption{Layer-wise task specific redudancy for SEM task. Redundant layers are represented by the colored blocks.}
	\label{fig:layer-redundancy-sem}
\end{figure*}

\begin{figure}[t]
	\begin{subfigure}{.98\linewidth}
		\centering
		\includegraphics[width=\linewidth]{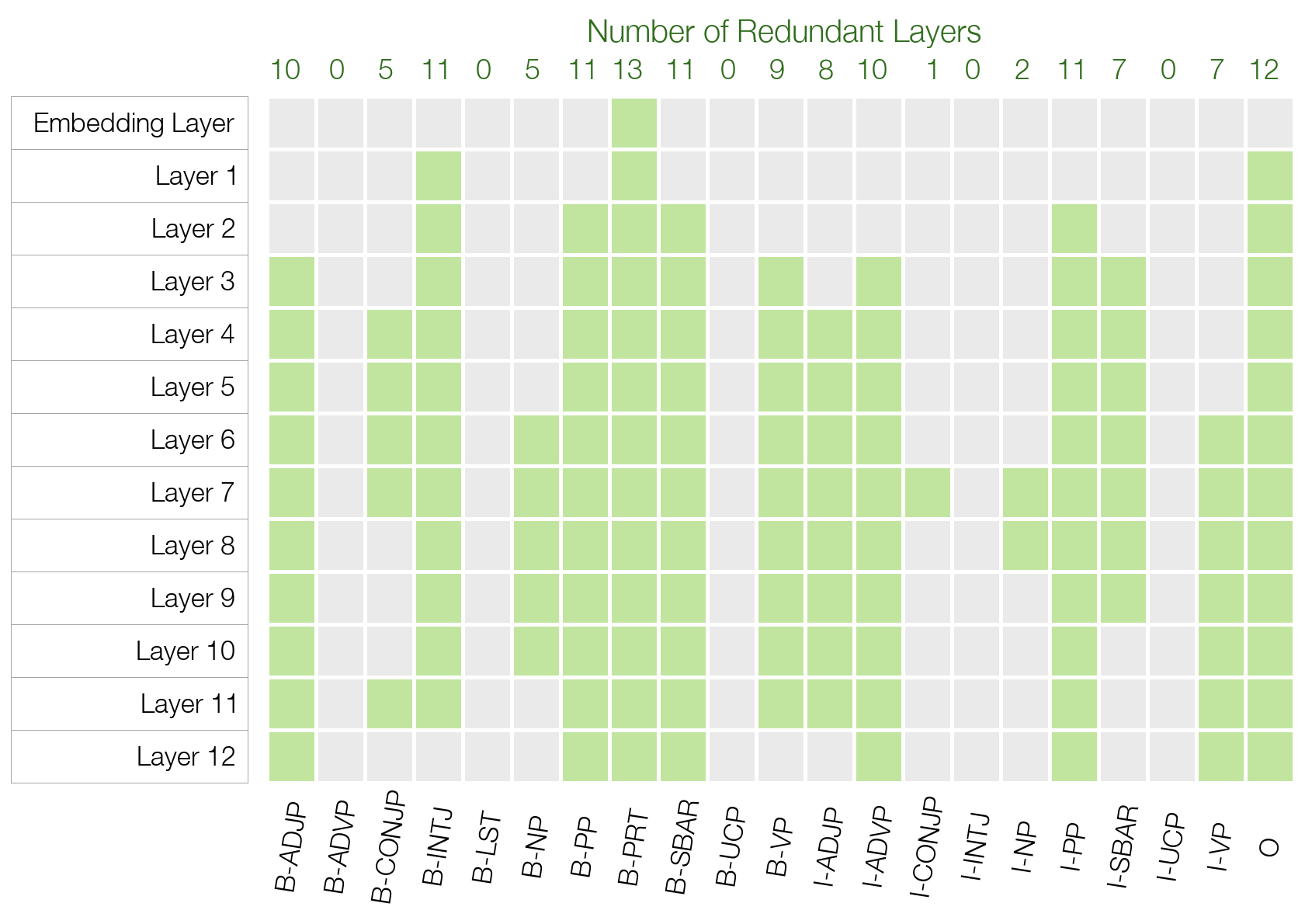}  
		\caption{BERT}
		\label{subfig:bert-layer-redundancy-chunking}
	\end{subfigure}
	\begin{subfigure}{.98\linewidth}
		\centering
		\includegraphics[width=\linewidth]{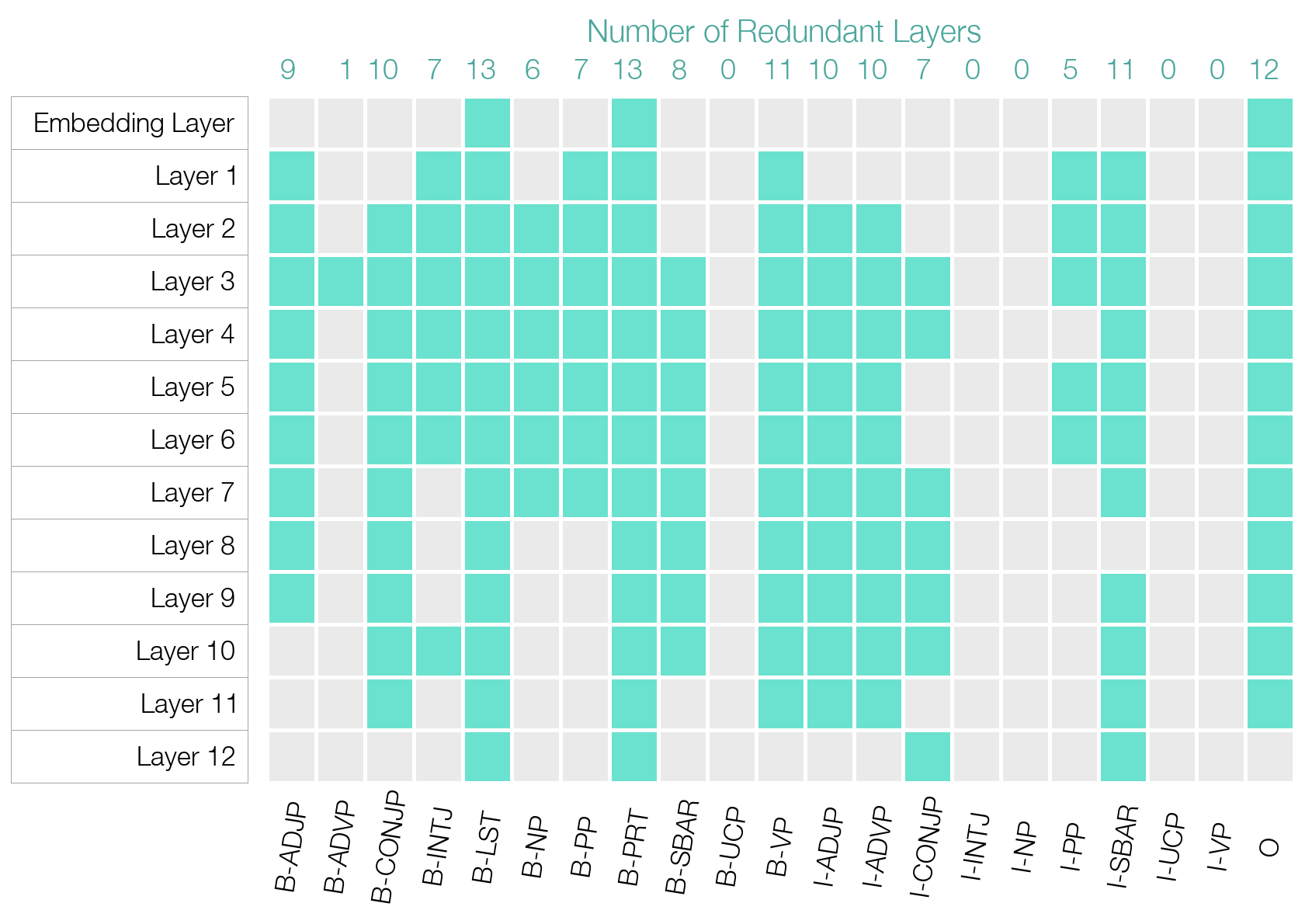}
		\caption{XLNet}
		\label{subfig:xlnet-layer-redundancy-chunking}
	\end{subfigure}
	\caption{Layer-wise task specific redudancy for Chunking task. Redundant layers are represented by the colored blocks.}
	\label{fig:layer-redundancy-chunking}
\end{figure}

	Figures \ref{fig:layer-redundancy-pos}, \ref{fig:layer-redundancy-sem} and \ref{fig:layer-redundancy-chunking} show the layer-wise task-specific redundancy for individual classes within POS, SEM and Chunking respectively. We do not present these fine-grained plots for CCG (over 1000 classes) or sequence classification tasks (binary classification only).
	
	\begin{table*}[t]
		\begin{subtable}{0.99\linewidth}
			\centering
			\resizebox{\linewidth}{!}{
				\begin{tabular}{r|cccc|ccccccc}
					\toprule
					& POS & SEM & CCG & Chunking & SST-2 & MRPC & MNLI & QNLI & QQP & RTE & STS-B \\
					\midrule
					Oracle & 95.2\% & 92.0\% & 90.1\% & 94.6\% & 90.6\% & 86.0\% & 81.7\% & 90.2\% & 91.2\% & 69.3\% & 89.7\% \\
					1\% Loss & 94.2\% & 91.1\% & 89.2\% & 93.6\% & 89.7\% & 85.2\% & 80.9\% & 89.3\% & 90.2\% & 68.6\% & 88.8\% \\
					\midrule
					Embedding & 89.6\% & 81.5\% & 70.0\% & 77.5\% & 50.9\% & 68.4\% & 31.8\% & 49.5\% & 63.2\% & 52.7\% & 0.0\% \\
					Layer 1 & 93.1\% & 87.6\% & 78.9\% & 82.1\% & 78.4\% & 68.9\% & 42.8\% & 59.7\% & 71.4\% & 52.7\% & 6.0\% \\
					Layer 2 & \textbf{95.3\%} & \textbf{91.7\%} & 86.6\% & 91.0\% & 80.2\% & 71.3\% & 45.0\% & 61.2\% & 73.3\% & 56.0\% & 10.4\% \\
					Layer 3 & \textbf{95.5\%} & \textbf{92.3\%} & 88.0\% & 92.0\% & 80.6\% & 69.6\% & 54.0\% & 74.4\% & 77.2\% & 54.9\% & 54.5\% \\
					Layer 4 & \textbf{96.0\%} & \textbf{93.0\%} & \textbf{89.6\%} & \textbf{94.0\%} & 81.2\% & 75.5\% & 61.8\% & 81.3\% & 80.1\% & 55.6\% & 84.9\% \\
					Layer 5 & \textbf{96.0\%} & \textbf{93.2\%} & \textbf{90.4\%} & \textbf{94.0\%} & 82.3\% & 76.2\% & 65.9\% & 82.9\% & 84.4\% & 59.6\% & 85.8\% \\
					Layer 6 & \textbf{96.3\%} & \textbf{93.4\%} & \textbf{91.6\%} & \textbf{94.9\%} & 86.2\% & 77.5\% & 71.6\% & 83.2\% & 85.8\% & 62.1\% & 86.4\% \\
					Layer 7 & \textbf{96.2\%} & \textbf{93.3\%} & \textbf{91.9\%} & \textbf{95.1\%} & 88.6\% & 79.4\% & 74.9\% & 83.8\% & 86.9\% & 62.5\% & 86.8\% \\
					Layer 8 & \textbf{96.0\%} & \textbf{93.1\%} & \textbf{91.9\%} & \textbf{94.8\%} & \textbf{90.6\%} & 77.5\% & 76.4\% & 84.4\% & 87.1\% & 63.5\% & 87.1\% \\
					Layer 9 & \textbf{95.8\%} & \textbf{92.9\%} & \textbf{91.6\%} & \textbf{94.5\%} & \textbf{90.5\%} & 83.3\% & 79.8\% & 84.8\% & 87.7\% & 63.2\% & 87.0\% \\
					Layer 10 & \textbf{95.6\%} & \textbf{92.5\%} & \textbf{91.2\%} & \textbf{94.1\%} & \textbf{90.6\%} & 82.6\% & 80.3\% & 86.1\% & 89.0\% & 64.3\% & 87.3\% \\
					Layer 11 & \textbf{95.4\%} & \textbf{92.3\%} & \textbf{90.9\%} & \textbf{93.9\%} & \textbf{90.4\%} & \textbf{85.8\%} & \textbf{81.7\%} & \textbf{89.8\%} & \textbf{91.0\%} & 66.4\% & \textbf{88.9\%} \\
					Layer 12 & \textbf{95.1\%} & \textbf{92.0\%} & \textbf{90.2\%} & 93.2\% & \textbf{90.1\%} & \textbf{87.3\%} & \textbf{82.0\%} & \textbf{90.4\%} & \textbf{91.1\%} & 66.1\% & \textbf{89.7\%} \\
					\bottomrule
				\end{tabular}
			}
			\caption{BERT}
			\label{tab:full_results_task_specific_layer_bert}
		\end{subtable} \\
		\\
		\begin{subtable}{0.99\linewidth}
			\centering
			\resizebox{\linewidth}{!}{
				\begin{tabular}{r|cccc|ccccccc}
					\toprule
					& POS & SEM & CCG & Chunking & SST-2 & MRPC & MNLI & QNLI & QQP & RTE & STS-B \\
					\midrule
					Oracle & 95.9\% & 92.5\% & 90.8\% & 94.2\% & 92.4\% & 86.5\% & 78.9\% & 88.7\% & 87.2\% & 71.1\% & 88.9\% \\
					1\% Loss & 95.0\% & 91.5\% & 89.9\% & 93.3\% & 91.5\% & 85.7\% & 78.1\% & 87.8\% & 86.4\% & 70.4\% & 88.0\% \\
					\midrule
					Embedding & 89.5\% & 82.6\% & 70.5\% & 77.0\% & 50.9\% & 68.4\% & 32.7\% & 50.5\% & 63.2\% & 52.7\% & 0.6\% \\
					Layer 1 & \textbf{96.3\%} & \textbf{92.9\%} & 88.7\% & 90.8\% & 79.6\% & 70.6\% & 44.2\% & 58.9\% & 72.0\% & 47.3\% & 8.8\% \\
					Layer 2 & \textbf{96.7\%} & \textbf{93.6\%} & \textbf{91.0\%} & \textbf{93.4\%} & 81.1\% & 70.1\% & 45.1\% & 58.6\% & 73.8\% & 45.8\% & 11.0\% \\
					Layer 3 & \textbf{96.8\%} & \textbf{93.5\%} & \textbf{91.8\%} & \textbf{94.2\%} & 84.7\% & 71.1\% & 61.6\% & 74.2\% & 82.4\% & 47.3\% & 81.1\% \\
					Layer 4 & \textbf{96.7\%} & \textbf{93.4\%} & \textbf{92.1\%} & \textbf{94.2\%} & 88.3\% & 76.0\% & 63.7\% & 74.1\% & 85.0\% & 53.1\% & 82.8\% \\
					Layer 5 & \textbf{96.6\%} & \textbf{93.2\%} & \textbf{92.4\%} & \textbf{93.9\%} & 88.6\% & 79.4\% & 68.4\% & 81.3\% & \textbf{89.2\%} & 62.1\% & 84.9\% \\
					Layer 6 & \textbf{96.3\%} & \textbf{92.6\%} & \textbf{92.0\%} & \textbf{94.2\%} & 90.1\% & 83.1\% & 73.9\% & 83.3\% & \textbf{89.9\%} & 63.5\% & 85.9\% \\
					Layer 7 & \textbf{96.1\%} & \textbf{92.3\%} & \textbf{91.9\%} & \textbf{94.0\%} & \textbf{92.9\%} & 85.3\% & \textbf{79.1\%} & \textbf{88.1\%} & \textbf{89.9\%} & 67.1\% & 86.7\% \\
					Layer 8 & \textbf{95.8\%} & \textbf{91.9\%} & \textbf{91.6\%} & \textbf{93.5\%} & \textbf{93.6\%} & \textbf{87.7\%} & \textbf{80.7\%} & \textbf{90.0\%} & \textbf{89.2\%} & 65.0\% & 87.6\% \\
					Layer 9 & \textbf{95.3\%} & \textbf{91.6\%} & \textbf{91.4\%} & 93.1\% & \textbf{94.2\%} & \textbf{87.5\%} & \textbf{80.1\%} & \textbf{90.3\%} & \textbf{88.4\%} & 69.3\% & \textbf{88.2\%} \\
					Layer 10 & 94.9\% & 91.2\% & \textbf{90.8\%} & 92.1\% & \textbf{93.8\%} & \textbf{86.5\%} & \textbf{80.1\%} & \textbf{90.4\%} & \textbf{88.9\%} & \textbf{71.8\%} & \textbf{88.2\%} \\
					Layer 11 & 94.6\% & 90.8\% & \textbf{90.2\%} & 91.1\% & \textbf{94.5\%} & \textbf{86.8\%} & \textbf{80.1\%} & \textbf{90.5\%} & \textbf{88.5\%} & \textbf{71.8\%} & \textbf{88.5\%} \\
					Layer 12 & 92.0\% & 87.4\% & 86.0\% & 85.9\% & \textbf{93.8\%} & \textbf{86.5\%} & \textbf{80.8\%} & \textbf{90.6\%} & \textbf{89.3\%} & \textbf{71.1\%} & \textbf{88.5\%} \\
					\bottomrule
				\end{tabular}
			}
			\caption{XLNet}
			\label{tab:full_results_task_specific_layer_xlnet}
		\end{subtable}
		\caption{Task specific layer wise results across all tasks. The oracle is trained on all 13 layers combined. Bold numbers highlight layers for each task that maintain 99\% of the Oracle's performance}
		\label{tab:full_results_task_specific_layer}
	\end{table*}
	
	\subsection{Task-Specific Neuron-level Redundancy}
	Tables \ref{subtab:appendix-task-specific-neuron-bert} and \ref{subtab:appendix-task-specific-neuron-xlnet} provide the per-task detailed results along with reduced accuracies after running task-specific neuron-level redundancy analysis.
	
	\begin{table*}[t]
		\begin{subtable}{0.49\linewidth}
			\centering
			\resizebox{\linewidth}{!}{
				\begin{tabular}{r|cccc|ccccccc|c}
					\toprule
					Task & Oracle & \#Neurons & Reduced Accuracy \\
					\midrule
					POS & 95.7\% & 290 & 94.3\% \\
					SEM & 92.2\% & 330 & 90.8\% \\
					CCG & 89.9\% & 330 & 88.7\% \\
					Chunking & 94.4\% & 750 & 93.8\% \\
					\midrule
					Word Average & 93.1\% & 425 & 91.9\% \\
					\midrule
					SST-2 & 90.6\% & 30 & 88.4\% \\
					MRPC & 86.3\% & 190 & 85.0\% \\
					MNLI & 81.7\% & 30 & 81.8\% \\
					QNLI & 90.3\% & 40 & 89.1\% \\
					QQP & 91.2\% & 10 & 90.8\% \\
					RTE & 69.7\% & 320 & 68.6\% \\
					STS-B & 89.6\% & 290 & 88.3\% \\
					\midrule
					Sentence Average & 85.6\% & 130 & 84.6\% \\
					\bottomrule
				\end{tabular}
			}
			\caption{BERT}
			\label{subtab:appendix-task-specific-neuron-bert}
		\end{subtable}%
		\begin{subtable}{0.49\linewidth}
			\centering
			\resizebox{\linewidth}{!}{
				\begin{tabular}{r|cccc|ccccccc|c}
					\toprule
					Task & Oracle & \#Neurons & Reduced Accuracy \\
					\midrule
					POS & 96.1\% & 280 & 95.6\% \\
					SEM & 92.2\% & 290 & 91.1\% \\
					CCG & 90.2\% & 690 & 89.8\% \\
					Chunking & 94.1\% & 660 & 93.0\% \\
					\midrule
					Word Average & 93.2\% & 480 & 92.4\% \\
					\midrule
					SST-2 & 92.9\% & 70 & 91.3\% \\
					MRPC & 85.8\% & 170 & 85.0\% \\
					MNLI & 79.0\% & 90 & 77.9\% \\
					QNLI & 88.3\% & 20 & 88.5\% \\
					QQP & 87.4\% & 20 & 88.0\% \\
					RTE & 70.4\% & 400 & 71.1\% \\
					STS-B & 88.9\% & 300 & 86.6\% \\
					\midrule
					Sentence Average & 84.7\% & 152 & 84.1\% \\
					\bottomrule
				\end{tabular}
			}
			\caption{XLNet}
			\label{subtab:appendix-task-specific-neuron-xlnet}
		\end{subtable}
		\caption{Accuracies after running \textit{linguistic correlation analysis} and extracting the minimal set of neurons from all 9984 neurons}
		\label{tab:appendix-task-specific-neuron}
	\end{table*}
	
	\subsection{Application: Efficient Feature Selection}
	\subsubsection{Transfer Learning Detailed Results}
	Tables \ref{tab:appendix-results_word} and \ref{tab:appendix-results_sentence} show the detailed per-task results for our proposed feature selection algorithm. 
	\label{subsec:appendix-transfer-learning}
	\begin{table}[t]
		\centering
		\resizebox{\columnwidth}{!}{
			\begin{tabular}{c|l|rrrrr}
				\toprule
				& & POS & SEM & CCG & Chunking  \\
				\cmidrule{1-6}
				\parbox[t]{2mm}{\multirow{7}{*}{\rotatebox[origin=c]{90}{BERT}}} & Oracle & 95.2\% & 92.0\% & 90.1\% & 94.6\% \\ 
				& Neurons & \multicolumn{4}{c}{9984} \\
				\cmidrule{2-6}
				& LS & 94.8\% & 91.2\% & 89.2\% & 94.0\%  \\ 
				& Layers & 3 & 3 & 8 & 7  \\
				\cmidrule{2-6}
				& \texttt{CCFS} & 93.9\% & 90.1\% & 90.2\% & 93.7\%  \\
				& Neurons & 300 & 400 & 400 & 600  \\
				\cmidrule{2-6}
				& \% Reduct.    & 97\%$\downarrow$ & 96\%$\downarrow$ & 96\%$\downarrow$ & 94\%$\downarrow$ \\
				\bottomrule
				\parbox[t]{2mm}{\multirow{6}{*}{\rotatebox[origin=c]{90}{XLNet}}} & Oracle & 95.9\% & 92.5\% & 90.8\% & 94.2\%  \\ 
				& Neurons & \multicolumn{4}{c}{9984} \\
				\cmidrule{2-6}
				& LS & 96.3\% & 92.9\% & 90.3\% & 93.5\%   \\ 
				& Layers & 2 & 2 & 3 & 3  \\
				\cmidrule{2-6}
				& \texttt{CCFS} & 95.6\% & 91.9\% & 89.5\% & 91.8\%  \\
				& Neurons & 300 & 400 & 300 & 600  \\
				\cmidrule{2-6}
				& \% Reduct.    & 97\%$\downarrow$ & 96\%$\downarrow$ & 97\%$\downarrow$ & 94\%$\downarrow$  \\
				\bottomrule
			\end{tabular}
		}
		\caption{Results of sequence labeling tasks using \texttt{LayerSelector}(\emph{LS}) with performance threshold$=1$ and \texttt{CCFS} with performance threshold$=1$. Oracle is using a concatenation of all layers. \emph{Layers} shows the number of the selected layer. \textit{Neurons} are the final number of neurons (features) used for classification. \emph{$\%$ Reduct.} shows the percentage reduction in neurons compared to the full network.}
		\label{tab:appendix-results_word}
	\end{table}
	
	\begin{table*}[t]
		\centering
		\begin{tabular}{c|l|rrrrrrr}
			\toprule
			& & SST-2 & MRPC & MNLI & QNLI & QQP & RTE & STS-B \\
			\cmidrule{1-9}
			\parbox[t]{2mm}{\multirow{7}{*}{\rotatebox[origin=c]{90}{BERT}}} & Oracle & 90.6\% & 86.0\% & 81.7\% & 90.2\% & 91.2\% & 69.3\% & 89.7\%   \\ 
			& Neurons & \multicolumn{7}{c}{9984} \\
			\cmidrule{2-9}
			& LS & 88.2\% & 86.0\% & 81.6\% & 89.9\% & 90.9\% & 69.3\% & 89.1\%\\
			& Layers & 8 & 12 & 12 & 12 & 12 & 13 & 12 \\
			\cmidrule{2-9}
			& \texttt{CCFS} & 87.0\% & 86.3\% & 81.3\% & 89.1\% & 89.9\% & 65.7\% & 88.6\%\\
			& Neurons & 30 & 100 & 30 & 10 & 20 & 30 & 400 \\
			\cmidrule{2-9}
			& \% Reduction & 99.7\%$\downarrow$ & 99.0\%$\downarrow$ & 99.7\%$\downarrow$ & 99.9\%$\downarrow$ & 99.8\%$\downarrow$ & 99.9\%$\downarrow$ & 96.0\%$\downarrow$ \\
			\bottomrule
			
			\parbox[t]{2mm}{\multirow{7}{*}{\rotatebox[origin=c]{90}{XLNet}}} & Oracle & 92.4\% & 86.5\% & 78.9\% & 88.7\% & 87.2\% & 71.1\% & 88.9\% \\ 
			& Neurons & \multicolumn{7}{c}{9984} \\
			\cmidrule{2-9}
			& LS & 88.2\% & 86.0\% & 79.9\% & 88.8\%  &  89.3\% & 71.1\% & 88.1\% \\ 
			& Layers & 6 & 9 & 8 & 8 & 6 & 11 & 9 \\
			\cmidrule{2-9}
			& \texttt{CCFS} & 87.5\% & 89.0\% & 78.4\% & 88.3\% & 88.8\% & 69.0\% & 87.2\%  \\
			& Neurons & 50 & 100 & 50 & 200 & 100 & 100 & 400 \\
			\cmidrule{2-9}
			& \% Reduction & 99.5\%$\downarrow$ & 99.0\%$\downarrow$ & 99.5\%$\downarrow$ & 98.0\%$\downarrow$ & 99.0\%$\downarrow$ & 99.0\%$\downarrow$ & 96.0\%$\downarrow$ \\
			\bottomrule
		\end{tabular}
		\caption{Results of sequence classification tasks using \texttt{LayerSelector}(\emph{LS}) with performance threshold$=1$ and \texttt{CCFS} with performance threshold$=1$. Oracle is using a concatenation of all layers. \emph{Layers} shows the number of the selected layer. \textit{Neurons} are the final number of neurons (features) used for classification. \emph{$\%$ Reduct.} shows the percentage reduction in neurons compared to the full network.}
		\label{tab:appendix-results_sentence}
	\end{table*}
	
	\subsubsection{Pretrained model timing analysis}
	\label{subsec:appendix-pretrained-timing}
\begin{figure}[ht]
		\centering
		\includegraphics[width=\linewidth]{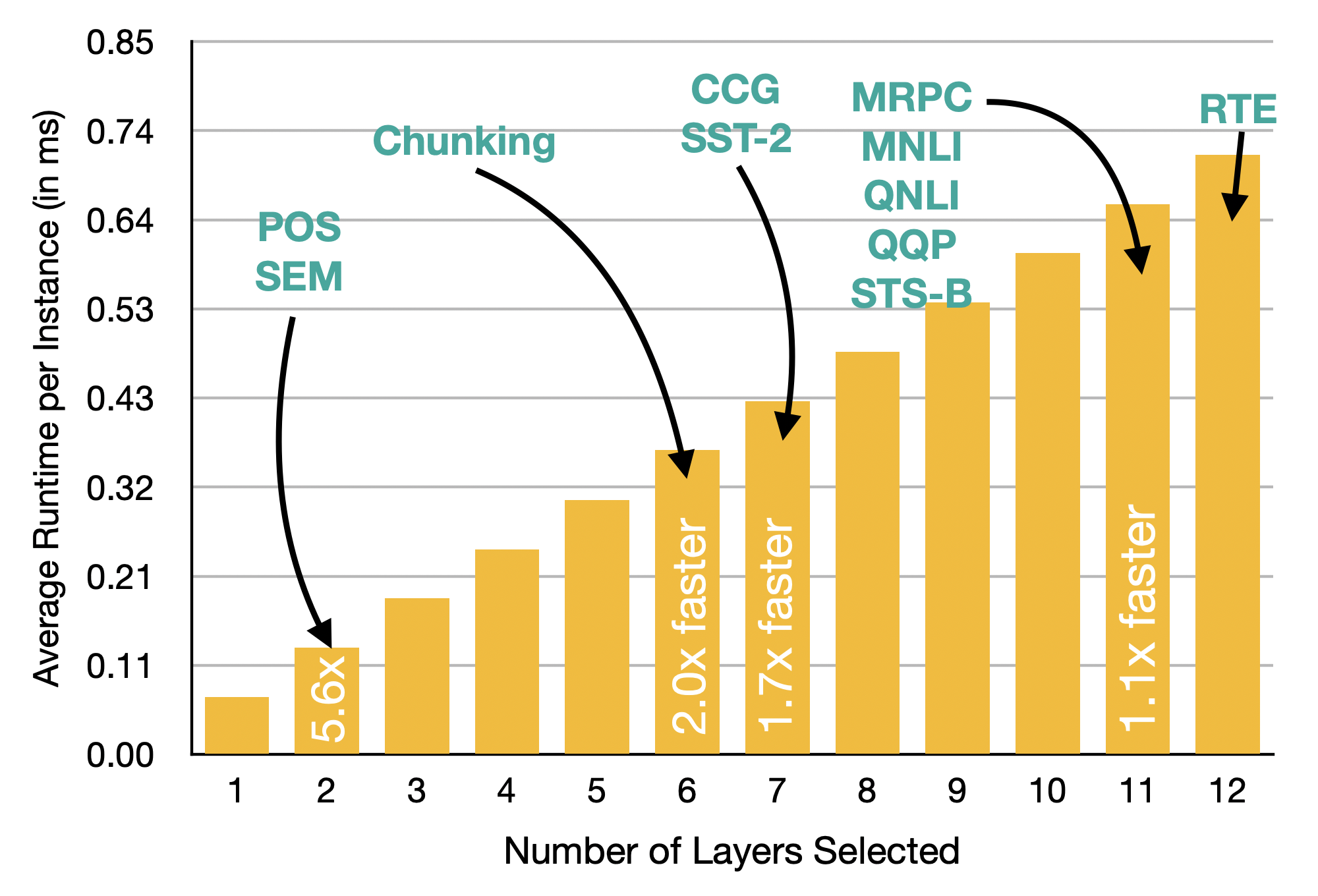}
		\caption{Average runtime per instance computed across all sequence classification tasks for BERT. Sequence classification tasks all have a near $2{\tt x}$ speed up, while most sequence labeling tasks have a $1.08{\tt x}$ speedup. }
		\label{fig:appendix-bert_timing}
	\end{figure}

\begin{figure}[ht]
		\centering
		\includegraphics[width=\linewidth]{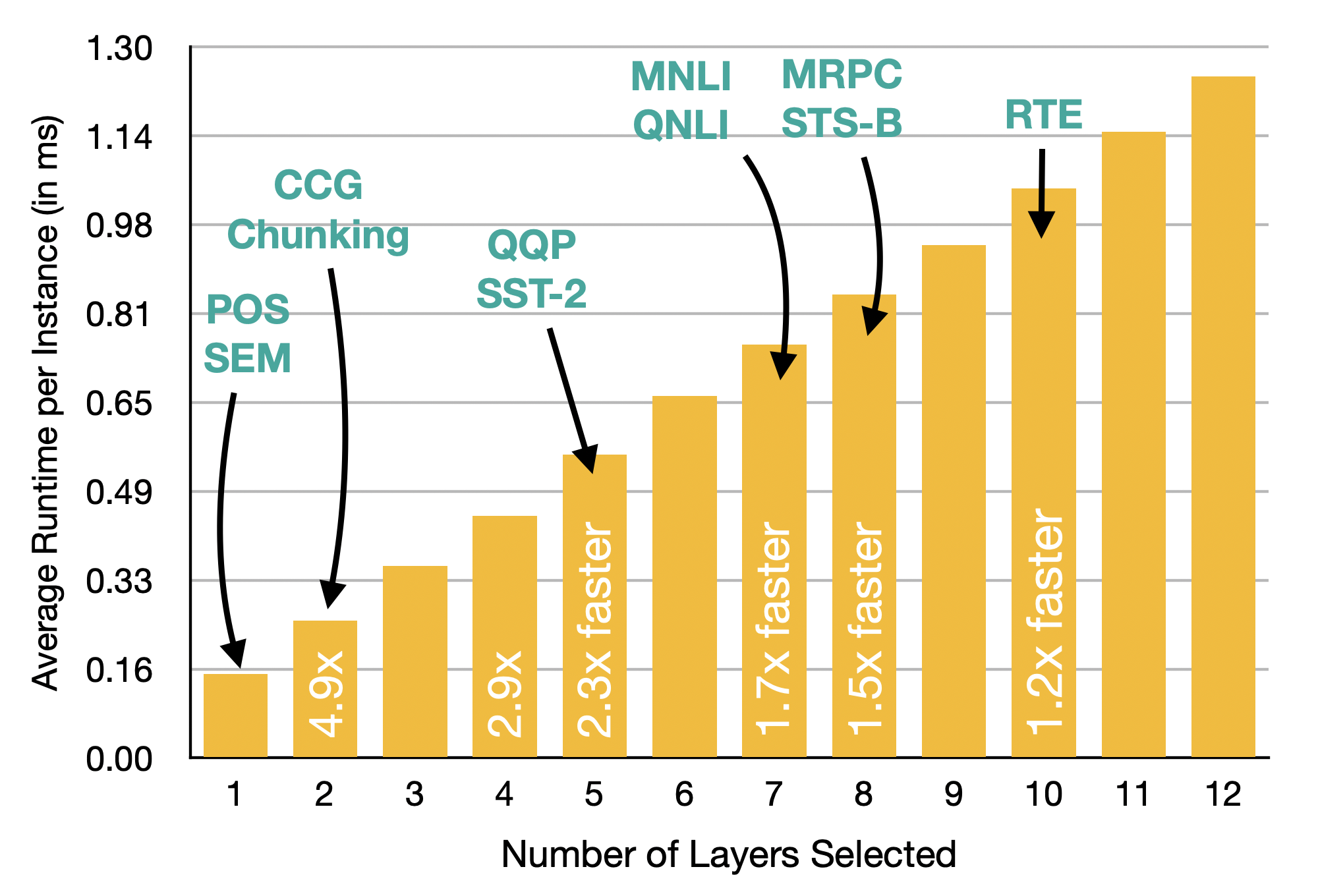}
		\caption{Average runtime per instance computed across all sequence classification tasks for XLNet. Sequence classification tasks all have a near $2{\tt x}$ speed up, while most sequence labeling tasks have a $1.08{\tt x}$ speedup. }
		\label{fig:appendix-xlnet_timing}
	\end{figure}

	The average runtime per instance was computed by dividing the total number of seconds taken to run the forward pass for all batches by the total number of sentences. All computation was done on an NVidia GeForce GTX TITAN X, and the numbers are averaged across 3 runs. Figures \ref{fig:appendix-bert_timing} and \ref{fig:appendix-xlnet_timing} shows the results of various number of layers (with the selected layer highlighted for each task). 
	\label{sec:appendix}

    	\subsubsection{Training time analysis}
	\label{subsec:appendix-training-time}
\begin{figure}[ht]
		\centering
		\includegraphics[width=\linewidth]{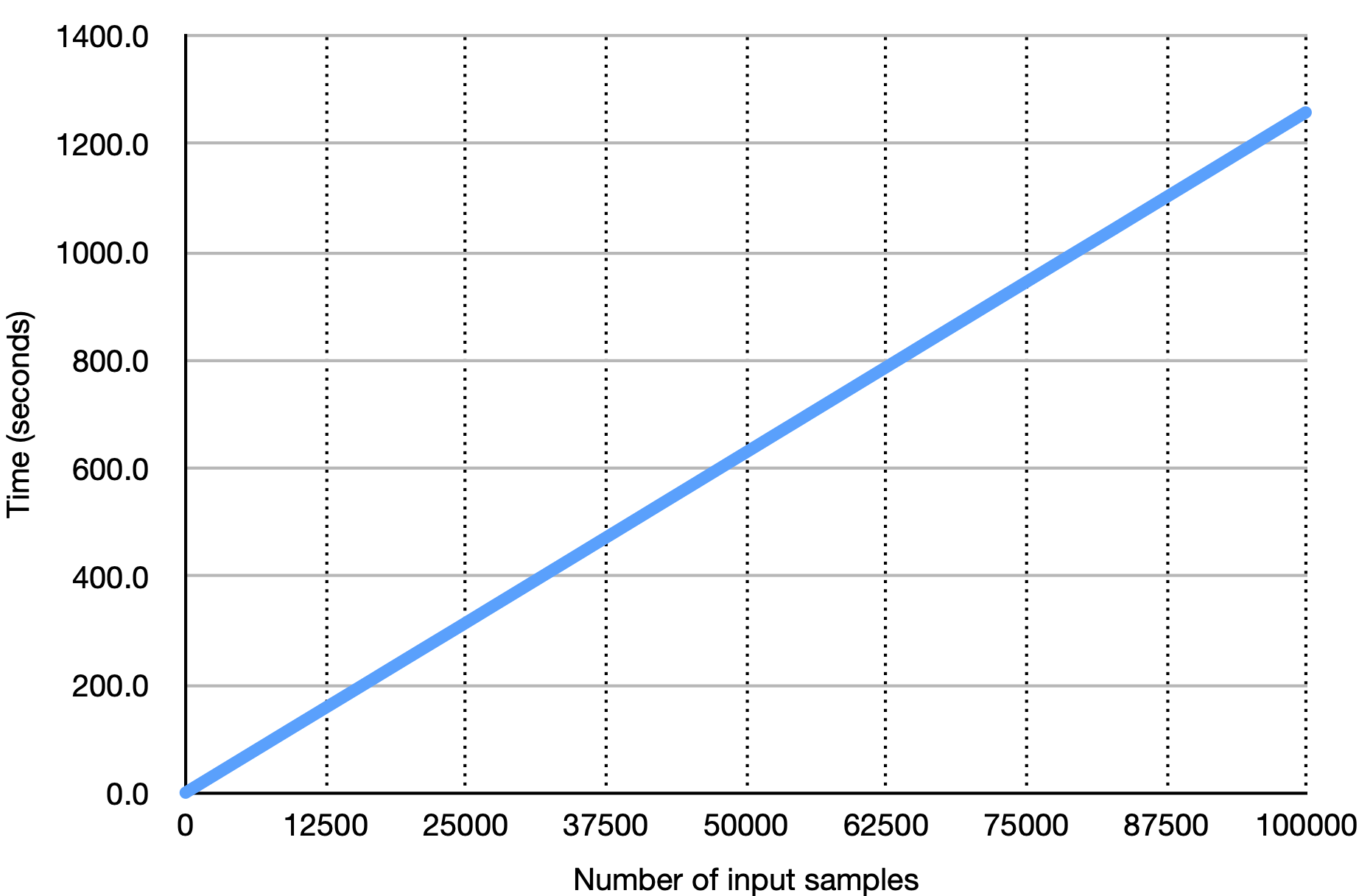}
		\caption{Runtime vs number of examples when extracting contextual embeddings using BERT}
		\label{fig:training-time-bert}
	\end{figure}

\begin{figure}[ht]
	\centering
		\includegraphics[width=\linewidth]{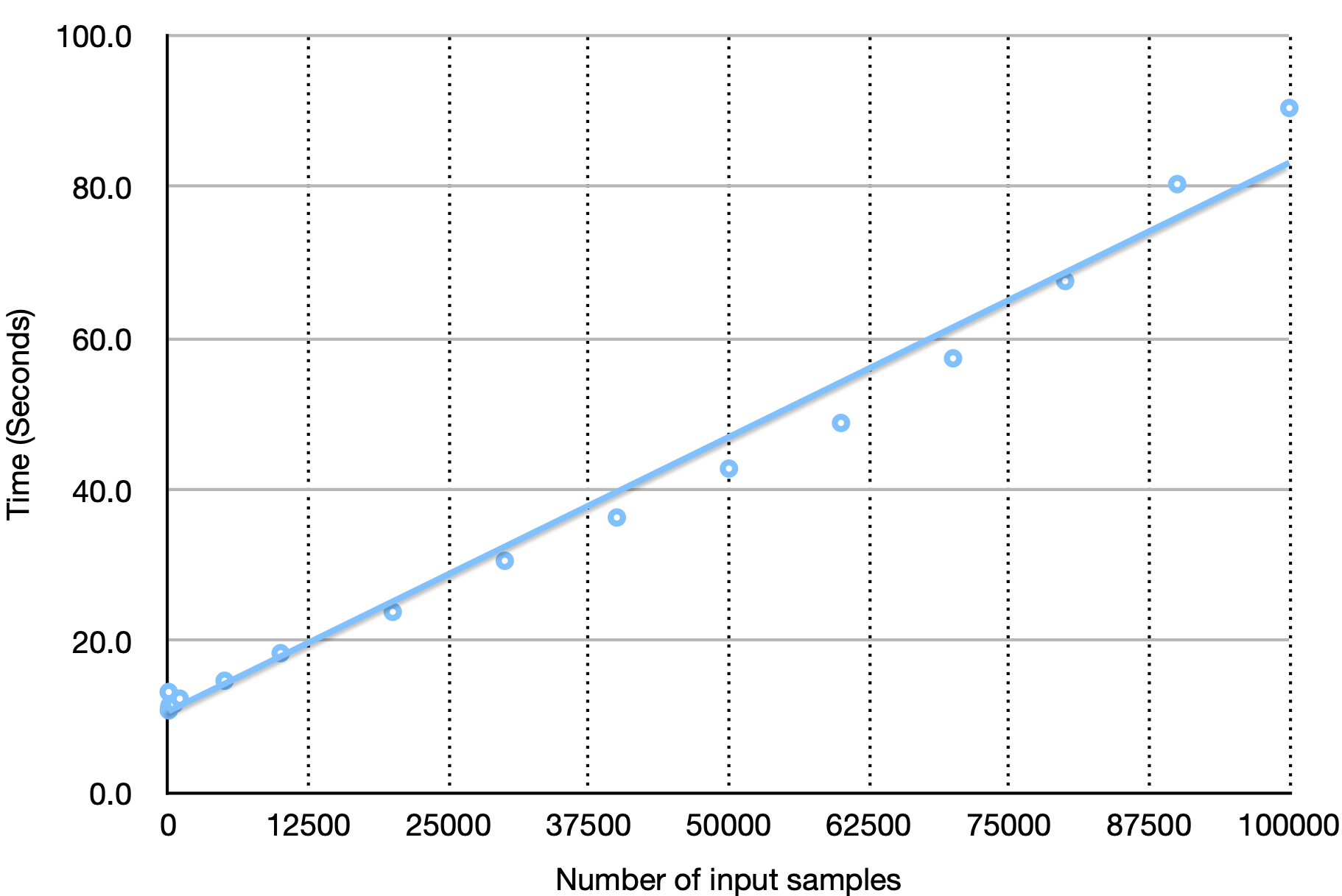}
	\caption{Runtime vs number of examples when performing correlation clustering}
	\label{fig:training-time-cc}
\end{figure}

\begin{figure}[ht]
	\centering
		\includegraphics[width=\linewidth]{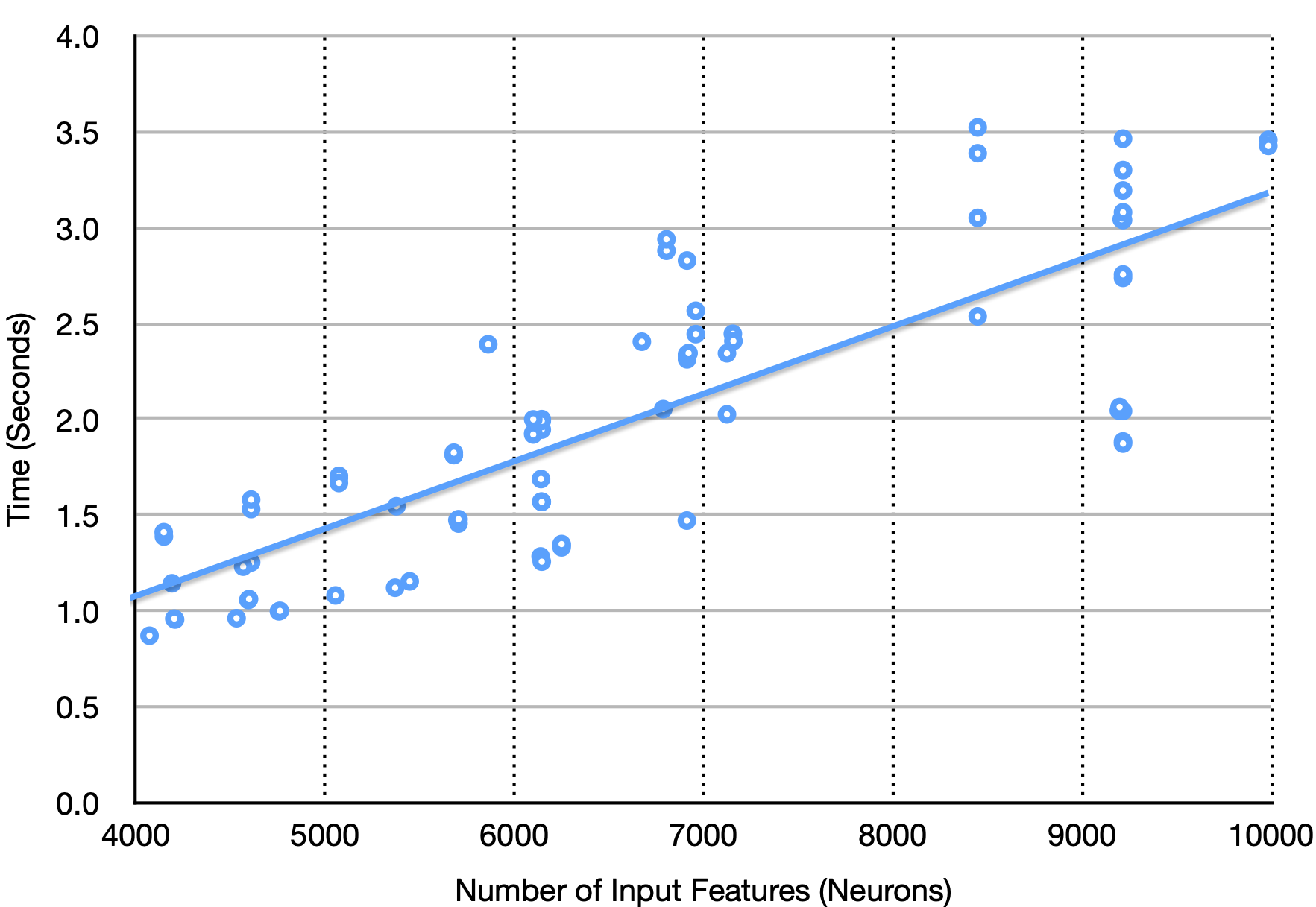}
	\caption{Runtime vs number of features when performing feature ranking using the weights of a trained classifier}
	\label{fig:training-time-ranking}
\end{figure}

	Figures \ref{fig:training-time-bert}, \ref{fig:training-time-cc} and \ref{fig:training-time-ranking} show the runtimes of the various steps of the proposed efficient feature selection for transfer learning application. Extraction of features and correlation clustering both scale linearly as the number of input tokens increases, while ranking the various features scales linearly with the number of total features.
	

	\subsection{Center Kernel Alignment}
	\label{subsec:appendix-cka}
For layer-level redundancy, we compare representations from various layers using linear Center Kernel Alignment (\texttt{cka} - \newcite{pmlr-v97-kornblith19a}). Here, we briefly present the mathematical definitions behind \texttt{cka}. Let $\mathbf{Z}$ denote a column centering transformation. As denoted in the paper, $\vz_j^i$ represents the contextualized embedding for some word $w_j$ at some layer $l_i$. Let $\vz^i$ represent the contextual embeddings over all $T$ words, i.e. it is of size $T \times N$ (where $N$ is the total number of neurons). Given two layers $x$ and $y$,
\begin{align*}
	\mathbf{X}, \mathbf{Y} = \mathbf{Z}\vz^x, \mathbf{Z}\vz^y
\end{align*}
the CKA similarity is
\begin{align*}
	{\tt cka}(\vz^x, \vz^y)
	:= \frac{\|\mathbf{X}^T\mathbf{Y}\|^2}{\|\mathbf{X}^T\mathbf{X}\|\|\mathbf{Y}^T\mathbf{Y}\|}
\end{align*}
where $\|\cdot\| $ is the Frobenius norm.

	%\section{Supplemental Material}
	%\label{sec:supplemental}
	%Submissions may include non-readable supplementary material used in the work and described in the paper.
	%Any accompanying software and/or data should include licenses and documentation of research review as appropriate.
	%Supplementary material may report preprocessing decisions, model parameters, and other details necessary for the replication of the experiments reported in the paper.
	%Seemingly small preprocessing decisions can sometimes make a large difference in performance, so it is crucial to record such decisions to precisely characterize state-of-the-art methods. 
	
\end{document}

%% file: math_commands.tex
\usepackage{amsmath,amsfonts,bm}

\def\eqref#1{equation~\ref{#1}}
\def\1{\bm{1}}

\def\vz{{\bm{z}}}

\DeclareMathAlphabet{\mathsfit}{\encodingdefault}{\sfdefault}{m}{sl}
\SetMathAlphabet{\mathsfit}{bold}{\encodingdefault}{\sfdefault}{bx}{n}

\def\sD{{\mathbb{D}}}